\documentclass{article}
\usepackage[T1]{fontenc}
\usepackage{iclr2027_conference,times}

\usepackage{amsmath,amsfonts,bm}

\def\eqref#1{equation~\ref{#1}}

\def\1{\bm{1}}

\DeclareMathAlphabet{\mathsfit}{\encodingdefault}{\sfdefault}{m}{sl}
\SetMathAlphabet{\mathsfit}{bold}{\encodingdefault}{\sfdefault}{bx}{n}

\usepackage{hyperref}
\usepackage{url}
\usepackage{graphicx}
\usepackage{booktabs}
\usepackage{xcolor}
\usepackage{enumitem}
\usepackage{rotating}
\usepackage{algorithm}
\usepackage{algpseudocode}
\usepackage{float}
\usepackage{placeins}

\newcommand{\HP}{[HP]}
\newcommand{\FP}{[FP]}
\newcommand{\figslot}[2][\linewidth]{%
  \IfFileExists{figures/#2}{\includegraphics[width=#1]{figures/#2}}{%
    \fbox{\parbox[c][2.6cm][c]{0.92\linewidth}{\centering\small\texttt{\detokenize{figures/#2}}\\
    placeholder: file not available locally}}}}
\newcommand{\tabslot}[1]{%
  \IfFileExists{tables/#1}{\input{tables/#1}}{%
    \fbox{\parbox[c][1.6cm][c]{0.92\linewidth}{\centering\small\texttt{\detokenize{tables/#1}}\\
    placeholder: file not available locally}}}}

\newcommand{\capFfour}{\textbf{Recoverability ceiling.} Cross-FSC half-bit crossing between the real 64 mT and 3T scans of the same subject (median over subjects, 95\% CI), mm [HP]: Zenodo T1w 3.16 (2.46--3.61, $n=8$), Zenodo T2w 3.19 (1.90--3.67, $n=10$), OpenNeuro T1w 4.81 (3.91--9.38, $n=11$). Dashed: end of the directly validated range, 7~mm [FP] (3.5~mm [HP]); values above it are coarser than the range the estimator was validated on.}

\newcommand{\capFGAP}{\textbf{Gap between real and synthetic degradation.} Per-subject half-bit crossings [HP] for the real 64 mT scan and the three noise-matched synthetics (one line per subject). Zenodo T1w gap to S-KSP+N (difference of median crossings) 1.16~mm [HP], above the 0.24~mm [HP] registration bound. Two OpenNeuro subjects sit at the grid cap (0.58~mm [HP]) and are shown as capped values.}

\newcommand{\capFUQ}{\textbf{Uncertainty controls} on the original registration frame, bands of 4~mm [FP] and coarser vs finer than 4~mm [FP]. Fine bands: AUROC 0.77, RHO\_own 0.39 vs RHO\_stranger 0.31; coarse bands: AUROC 0.68 (Table~\ref{tab:t3}). The validated-frame recompute is reported in Section~\ref{sec:models}.}

\newcommand{\capFSHARP}{\textbf{Apparent sharpness vs input ceiling}, 24 reference runs (marker = family) and SynthSR (star), by dataset column; horizontal lines mark each column's input ceiling (OpenNeuro T2w: registration-limited, qualitative, shading indicative only; NIST: no cross-field ceiling). Every trained run renders detail finer than its ceiling; SynthSR (5.2--6.1~mm [FP]) sits at it.}

\newcommand{\capTone}{\textbf{Master measurement table.} Every ceiling readout with its role, the synthetic gaps (difference of median crossings) with the registration bound, anisotropy, the NIST self-1FRC pair, and the direct-measurement control ladder. OpenNeuro T1w uses $n=11$ of 12 prepared subjects (one null crossing excluded).}

\newcommand{\capTtwo}{\textbf{Model audit.} One row per audited run (24) plus the external SynthSR model and per-family medians. Faithfulness is the merge scale of Algorithm~\ref{alg:rec} where computable; otherwise NC with a reason code (NC-EPI: NIST; NC-ANCHOR: Zenodo T2w; NC-XDEV: OpenNeuro T2w). LPIPS is tabulated for NIST runs; for the T1w diffusion runs it is compared against the raw input per subject in Section~\ref{sec:models}.}

\newcommand{\capTthree}{\textbf{Uncertainty controls per band (band centres in mm [FP]) and grouped} into bands of 4~mm [FP] and coarser versus finer than 4~mm [FP] (the split point used by the analysis; the whole-brain ceiling is 6.31~mm [FP]). Computed on the original registration frame; the validated-frame recompute is reported in Section~\ref{sec:models}. RHO: Spearman correlation between sample variance and band error against the own 3T (own), after partialling out band energy (partial), and against a stranger (stranger); AUROC: variance flagging top-decile band error.}

\newcommand{\capFINFO}{\textbf{Faithful information stops at the measured ceiling (display only).} One representative run per family on its own held-out subject, matched on the anterior lateral-ventricle margin: GAN on Zenodo T2w, ViT on NIST b0, diffusion on Zenodo T1w (posterior mean). Row 1: 64 mT input, true 3T, zoom ladder; row 2: model output at identical coordinates; last column: band-pass finer than the input ceiling (NIST, which has no cross-field ceiling: finer than the 64 mT self-resolution, 5.06~mm [FP]). Finer than the ceiling the input carries little subject information while every output renders fine detail; faithfulness is computed only for the diffusion row (Table~\ref{tab:t2}), and the GAN and ViT rows are not computable (NC-ANCHOR, NC-EPI).}

\newcommand{\capFSYNTH}{\textbf{Synthetic degradations keep a band the real acquisition lacks (display only).} Zenodo T1w sub-0048: true 3T, real 64 mT, and noiseless S-KSP, S-GAU, S-BIC, with zooms and a 4.0--6.3~mm [FP] band-pass between the synthetic and real crossings. Annotations are the noise-matched crossings: real 6.31 [FP], S-KSP+N 3.99, S-GAU+N 4.44, S-BIC+N 4.47.}

\newcommand{\capFEone}{\textbf{Cross-FSC curves.} Median over subjects of FSC against the 3T for the real 64 mT scan and the noise-matched synthetics (noiseless S-KSP dashed; it shares the 3T noise, so its curve is inflated), with the half-bit threshold, the empirical null band, and the 3T self-split scalar (2.47~mm [FP]).}

\newcommand{\capFQUALB}{\textbf{Confident detail across audit arms (display only).} Input and output with zooms for five arms, no truth column: GAN Zenodo T2w (NC-ANCHOR), GAN NIST b0 (NC-EPI), ViT NIST b0 (NC-EPI), multicontrast Zenodo T2w (NC-ANCHOR), single-contrast diffusion Zenodo T1w (faithfulness computed, merges 16 and 24~mm [FP], Table~\ref{tab:ladder}). The OpenNeuro T2w GAN arm is omitted here (outputs collapse under the cross-device gap, NC-XDEV) and reported in Table~\ref{tab:t2}.}

\title{You Cannot Recover What Was Never Measured: Quantifying the Information Ceiling of Ultra-Low-Field MRI
Super-Resolution}
\author{Prathamesh Pradeep Khole\textsuperscript{1}\thanks{Corresponding author: \texttt{pkhole@ucsc.edu}} \quad
Shreya Handa\textsuperscript{1,2} \quad
Utkarsh Gupta\textsuperscript{1} \quad
Razvan Marinescu\textsuperscript{1} \\
\textsuperscript{1}University of California, Santa Cruz \\
\textsuperscript{2}Johns Hopkins University}

\usepackage{graphicx}
\usepackage{booktabs}
\usepackage{amsmath}

\iclrfinalcopy

\begin{document}
\maketitle
\lhead{Preprint}

\begin{abstract}
Generative super-resolution models can turn portable 64 mT MRI into images that look like 3T scans, and the field evaluates them with PSNR, SSIM, and pixelwise uncertainty, most often on pairs built by synthetically degrading high-field images. Prior work acknowledges that these models hallucinate and that the problem is ill posed, but to our knowledge no study measures how much information about the individual subject the real low-field scan actually contains. We measure it. Using paired 64 mT and 3T scans of the same subjects from three public datasets, and a measurement protocol validated on tests whose correct answer is known in advance, we find that, judged over the whole brain, real 64 mT scans carry structure specific to the individual only down to approximately 3 to 4 mm half-pitch in plane, and coarser still through plane. Standard synthetic degradations preserve subject information roughly 1 mm beyond this ceiling, so models trained and benchmarked on synthetic pairs are evaluated on information that real scanners never record. We then test trained diffusion models and a publicly released external model on real paired acquisitions; 24 trained runs of five architectures (GAN, diffusion, and transformer families) give the coverage of the audit. On every subject where faithfulness can be measured, fine output detail is no more correlated with the subject's own 3T scan than with a stranger's, while sample-variance uncertainty does not distinguish fabricated structure from reconstruction difficulty. Because PSNR and SSIM score resemblance to a reference rather than whether detail belongs to the subject, a benchmark scored by them cannot tell recovery from fabrication. Code for the measurement protocol will be released so that recoverability claims can be tested for newer models.
\end{abstract}

\section{Introduction}
\label{sec:intro}

Portable 64 mT MRI reaches patients that fixed 3T systems cannot \citep{sheth2021portable}, and generative
super-resolution models are trained to map its images to 3T appearance \citep{iglesias2023lowfield}. Their
evaluation still rests on PSNR and SSIM \citep{wang2004ssim} against a high-field reference, most often on pairs
built by synthetically degrading high-field images, with a pixelwise uncertainty map as the safety signal. Each
of these tools measures how closely the output resembles a reference. None asks the question a clinician cares
about: does the fine structure in the output belong to this subject, or is it only a plausible sample from the
population?

The question has a physical answer. A reconstruction cannot contain more information about the subject than its
input does, so the finest scale at which a real 64 mT scan still carries structure specific to the individual
bounds what any model can recover from the input; subject-specific detail finer than that can come only from a
prior inferring it from coarser structure, which is a testable inference. Work on hallucination in inverse problems established that learned reconstructions
insert realistic false structure \citep{cohen2018distribution,antun2020instabilities,bhadra2021hallucinations},
and low-field super-resolution work acknowledges that detail generated where the input gives weak support is
unstable \citep{rediff}. These statements are qualitative: no study reports the bound in millimetres for a real
low-field scanner, so no benchmark can tell recovery from invention by the model.

We measure that number on paired 64 mT and 3T acquisitions of the same subjects from three public datasets, using a shared rigid-registration pipeline and estimators validated on synthetic cases with known ground truth.
We then test models trained on the real paired data against it, comparing each reconstruction with the true 3T
image of the same subject and with the 3T image of a stranger, once per spatial scale: the images are split into frequency bands from coarse to fine and the two comparisons are repeated in each band. Where the two correlations are equal, the output fits a stranger as well as the subject, so its detail at that scale is not specific to the subject. Two of these tools can be applied beyond MRI: cross-modality Fourier shell correlation (FSC) on paired real data as a shared-information ceiling, and a stranger-controlled, band-resolved faithfulness test
that reports, per scale, how much better an output matches its subject than a stranger.

Our contributions are the following.
\textbf{1)~A measured recoverability ceiling} (Claim~1; Section~\ref{sec:ceiling}): the FSC between the real 64 mT and 3T scans of the same subject (cross-FSC) crosses the half-bit threshold,
the standard FSC resolution criterion, at 3.16~mm half-pitch (\HP{}; conventions in
Section~\ref{sec:protocol}) on Zenodo T1w, and the limit is coarser through plane; to our knowledge this is the
first millimetre-scale measurement of that limit on a real low-field scanner.
\textbf{2)~A measured gap between synthetic and real degradation} (Claim~2; Section~\ref{sec:synthetic}): real
acquisition is coarser than the noise-matched k-space-truncation synthetic by 1.16~mm~\HP{},
so a synthetic benchmark scores recovery of content that a real scanner never records.
\textbf{3)~A direct fabrication measurement and an audit of trust tools} (Claim~3; Section~\ref{sec:models},
Tables~\ref{tab:ladder}, \ref{tab:t3} and~\ref{tab:t2}): on four held-out subjects, fine diffusion-output detail
is no more correlated with the subject's own 3T than with a stranger's, sample-variance uncertainty flags
reconstruction difficulty and not fabrication, and a public external model behaves the same way;
24 trained runs of five architectures give the coverage of the audit.
\textbf{4)~A measurement protocol} with its validation tests, algorithms, audit reason codes, and the approaches
that failed (Section~\ref{sec:protocol}, Appendices~\ref{app:alg} and~\ref{app:practice}).

\section{Related Work}
\label{sec:related}

\textbf{Low-field super-resolution calls the problem ill posed but never measures how ill posed.} Learned mappings from low-field or clinical-quality MRI to high-resolution appearance are trained mostly on pairs made by degrading high-field images \citep{iglesias2021synthsr,iglesias2023lowfield}. These models build on general-purpose GAN \citep{ledig2017srgan} and diffusion \citep{ho2020ddpm,saharia2022sr3} backbones, including MRI-specific diffusion variants \citep{safari2025ressrdiff}. Image Quality Transfer and related
methods synthesize low-field training inputs by blurring, downsampling, and contrast change
\citep{kalluvila2022,lin2023,figini2024}. A score on such inputs does not measure recovery from a real
acquisition. Clinical validation of low-field super-resolution has centred on morphometry
\citep{iglesias2023lowfield}, which operates at coarse scales consistent with the ceiling we measure; our
challenge is to fine-detail claims and to PSNR- and SSIM-scored benchmarks, not to morphometry. ReDiff \citep{rediff} observes that weak low-field support produces unstable high-frequency
generation and responds with reliability-guided sampling, but does not quantify the support at any spatial scale. A reader study reports a lesion-detectability threshold at 64 mT \citep{arnold2022}, a
task-performance number, distinct from the subject-information scale we measure. The OpenNeuro release
\citep{openneuro_ds006557} assesses test-retest reliability and the correspondence of 64 mT to 3T images through
regional morphometry and segmentation overlap; it does not estimate the spatial scale at which image
correspondence stops.

\textbf{Hallucination is documented, mostly without a scale.} Distribution-matching losses can add or remove
pathology \citep{cohen2018distribution}, learned reconstructions can insert false structure
\citep{antun2020instabilities}, and hallucination maps can be defined relative to a known forward operator's
null space \citep{bhadra2021hallucinations}. In low-field MRI the forward operator is not known in closed form,
so we measure the limit empirically from paired scans. PSNR, SSIM \citep{wang2004ssim}, and LPIPS
\citep{zhang2018lpips} trade off against each other \citep{blau2018perception}, but all score resemblance to a
reference, not whether detail belongs to the subject.

\textbf{Resolution and uncertainty tools.} Fourier shell correlation \citep{harauz1986fsc} with the half-bit
criterion \citep{vanheel2005halfbit}, Fourier ring correlation \citep{koho2019frc} usually compare two views of one acquisition, and decorrelation analysis
\citep{descloux2019decorr} reads resolution from a single image. We apply FSC between a low-field
and a high-field scan of the same subject, which turns a resolution measure into a measure of shared subject
information. Predictive variance \citep{kendall2017uncertainties} and calibrated intervals
\citep{angelopoulos2022im2im} are scored against the error of the output alone; we add a stranger control that asks whether the map tracks error for the correct subject or only regions that are hard to reconstruct in any brain.

\section{Measurement Protocol}
\label{sec:protocol}

\textbf{Conventions and preprocessing.} Half-pitch, \HP{}, is $1/(2k_c)$ and full-period, \FP{}, is $1/k_c$,
where $k_c$ is the cutoff spatial frequency. Intuitively, \FP{} is the width of one line pair (a bright band and the adjacent dark band) and \HP{} is the width of a single band, the smallest feature at that frequency; the two differ by a factor of two. For measurement, each volume is bias corrected with N4
\citep{tustison2010n4}, the 64 mT scan is registered to the 3T scan of the same subject with a rigid transform
under mutual information \citep{mattes2003mi}, and both are resampled once onto a common isotropic grid. Brain
and ventricle masks are SynthSeg labels \citep{billot2023synthseg} of the 3T scan carried by the known
transform. Deformable registration is excluded because a flexible warp could manufacture agreement between the two
scans that the acquisition did not record. (The reconstruction models of Section~\ref{sec:data} are trained on a separate 2D slice pipeline,
rigidly registered for Zenodo T1w and affinely registered for the others; every model output is mapped back
onto the rigid measurement grid by one interpolation.)

\textbf{Estimators and their roles.} The \emph{measurement} is cross-FSC (Algorithm~\ref{alg:fsc}): a gated 3D
Fourier shell correlation between the registered 64 mT and 3T volumes under a soft brain mask and Hann window \citep{harris1978windows}.
The reported scale is where the correlation falls below the half-bit threshold (the correlation at which the two images share about half a bit of information per Fourier component at that scale, \citealp{vanheel2005halfbit}) and stays below it for two
consecutive shells; an empirical null from subject-swapped pairs shows the correlation that shared anatomy and preprocessing produce without any subject-specific information. It is the only estimator that compares the two scans of one subject over the whole
brain. Two further single-image readouts, decorrelation of the 64 mT scan (Algorithm~\ref{alg:decorr}) and a blur sweep on
the 3T scan alone, are reported in Table~\ref{tab:t1} as consistency checks only; they are not independent
estimates of shared information and enter no claim.

\textbf{Known-answer gates.} Each estimator is run through pre-specified known-answer gates (A1 to A5, Table~\ref{tab:gates}), tests whose
correct answer is known in advance, before any value from it is used; the outcomes decide what role each
estimator plays. FSC recovers imposed cutoffs (gate A1) within 2.5--7.5\% over
3 to 6~mm~\FP{} but misses the 10\% criterion at 8~mm, so its validated range
ends near 7~mm~\FP{}; values coarser than that (the OpenNeuro T1w ceiling and several per-subject
anchors) lie beyond the directly validated range. The 3T self-split (A2) misses its 25\% criterion and is used only as
a fine upper bound, and decorrelation fails A1 and A5 (cutoff recovery and monotone response to added blur) and is kept as a qualitative check. Because FSC compares two different contrasts, its crossing could respond to the choice of intensity normalization rather than to shared information. Decision D1 therefore recomputes the real crossing under several intensity normalizations of the same images and compares the spread of those crossings with the real-versus-synthetic gap: the spread-to-gap ratio is 0.79 on Zenodo and 0.52 on OpenNeuro, below one on both, so the gap is not a normalization artifact and FSC stays primary.

\textbf{Synthetic degradations.} The 3T scan is degraded by k-space truncation (S-KSP), bicubic resampling
(S-BIC), or anisotropic Gaussian blur (S-GAU). A noiseless synthetic keeps the exact noise pattern of its
source image, so its agreement with that source is guaranteed by construction; the
comparison therefore uses noise-matched versions (S-KSP+N, S-BIC+N, S-GAU+N). S-GAU and S-BIC follow the
blur-and-resample recipes of Image Quality Transfer \citep{lin2023}, S-KSP is the physical band-limit model,
and randomized generators such as SynthSR sample over such operators rather than fixing one
\citep{iglesias2021synthsr}.
S-KSP zeroes every k-space frequency outside the Nyquist box of the native 64 mT voxel, so in-plane periods finer
than 3.2~mm~\FP{} and through-plane periods finer than 10~mm~\FP{} are
removed (the box is square, so its corners keep diagonal periods down to 2.26~mm); the
noise-matched versions add Rician noise at the level that reproduces the real 64 mT scan's signal-to-noise
ratio (mean brain intensity over background standard deviation). The gap is the difference between the median
real and the median synthetic crossing. Injecting a
1~mm shift moves the crossing by 0.473~mm~\FP{}; this is the
registration bound a gap must exceed.

\textbf{Faithfulness by scale (Algorithm~\ref{alg:rec}).} The idea is a stranger test: if the fine detail a
model outputs is no more correlated with the subject's own 3T scan than with a different person's, then that detail
cannot be about the subject. For each held-out subject and band $s$, REC($s$) is the correlation between the model output and the subject's
own 3T scan, SWAP($s$) is the correlation between the same output and the 3T scan of a stranger (the other
held-out subject of the same dataset, fixed before scoring and rigidly aligned), and CON($s$) is the mean
correlation between posterior samples. Every correlation in this readout is the
Pearson correlation between band-passed images, computed over the pooled 3T foreground voxels (intensity above
5\% of the maximum) of the sampled slices within the model's coverage
(Appendix~\ref{app:reconstruction-details}). Bands are one-octave annuli applied to each 2D slice with hard
edges; the 13 band centres from 32 to 2~mm~\FP{} sit 1.14 to 1.5 times apart, so adjacent bands
overlap and a cutoff spreads over its neighbours. The \emph{merge scale} is the coarsest band at which REC stays
within 0.05 of SWAP for two consecutive bands; because adjacent bands overlap, the persistence rule
guards against a single-band accident rather than testing two independent scales; the rule skips the coarsest
band, so 24~mm~\FP{} is the coarsest merge it can report. A known-answer test calibrates the
readout (Appendix~\ref{app:gates}): when a stranger's 3T replaces the subject's own at every scale finer than a
known cutoff, the recovered merge equals the cutoff or lies one grid step finer (two steps,
1.0~mm, at a 4~mm cutoff), with or without added noise and never coarser, so the readout can
only credit an output with slightly finer faithfulness than it has. Throughout the paper,
\emph{faithfulness} means this merge scale: detail finer than it is no more correlated with the subject than with a stranger; the per-band difference
REC minus SWAP is the quantity we suggest others report. The
rule assumes SWAP is non-negative; where SWAP is negative, a REC near zero can exceed the threshold without any
subject-specific agreement (Section~\ref{sec:models} reports one such case, and Table~\ref{tab:ladder} gives every
merge under a SWAP clipped at zero).
In any band, high CON with REC at SWAP is \emph{confident fabrication} in that band: the samples agree with one another but not with the subject. Two controls pass through the identical readout: the
registered real 64 mT input (the \emph{raw-input benchmark}) and a noise-matched k-space oracle built from the
subject's own 3T (S-KSP+N). The data-processing inequality bounds information, not estimator correlation: a
model may modestly exceed the raw input's band correlation at coarse, noise-swamped scales by denoising
structure the input carries, so a merge one band finer than the raw-input benchmark is not by itself information
creation. Each subject is read on the registration frame selected by the raw-input control alone, the frame on
which the registered input reproduces its anchor; that choice does not involve the model outputs
(Appendix~\ref{app:practice}).

\textbf{Uncertainty and audit.} Four controls test what per-voxel sample variance (VAR) responds to: its
Spearman correlation with the posterior-mean error against the own 3T (RHO\_own), the same after removing
(partialling out) the effect of band energy, its correlation with the error against a stranger (RHO\_stranger), and its voxel AUROC for
top-decile band error. Each audited run is scored only on its own held-out subjects after a contamination check (no held-out subject
appears in its training set). Four checks are applied to each held-out input: a 3T self-split fine bound (G-M1);
the input's cross-FSC crossing inside the CI of the dataset ceiling (G-M2, the input anchor); the input's
cross-FSC against the frozen stranger's 3T crossing at least one band coarser than the subject's own anchor, or
not at all (G-M3, so that the input tells its own 3T from a stranger's); and deterministic reproducibility
(G-M4). An arm enters the audit when at least two held-out subjects pass G-M2; for the direct arm every check is
reported per subject, and a subject that passes both G-M2 and G-M3 is called certified
(Section~\ref{sec:models}, Appendix~\ref{app:gates}). Where faithfulness cannot be computed a reason code
records why: NC-EPI (echo-planar distortion of the NIST low-field b0, the non-diffusion-weighted image), NC-ANCHOR (held-out subjects fail G-M2, Zenodo T2w), or
NC-XDEV (fewer than two held-out subjects pass under registration-limited cross-device correspondence, OpenNeuro
T2w). Every run
also receives apparent sharpness (decorrelation of the output alone, reported for coverage only because
decorrelation failed two gates), PSNR, and SSIM.

\section{Data and Audited Systems}
\label{sec:data}

\textbf{Datasets.} \emph{Zenodo} \citep{zenodo_paired} provides 64 mT and 3T T1-weighted (T1w) pairs ($n=8$
well-registered subjects in the T1w statistics) and T2-weighted (T2w) pairs ($n=10$), with a
$1.6\times1.6\times5.0$~mm low-field acquisition. \emph{OpenNeuro} ds006557
\citep{openneuro_ds006557} provides T1w ($n=11$; one subject with a null crossing is
excluded) and T2w pairs ($n=23$). \emph{NIST} \citep{nist_dualfield,nist_ir8580} provides
diffusion b0 images ($n=19/20$). These $n$ are measurement counts; the reconstruction
train/test partitions below are separate.

\textbf{Reference arm.} Proprietary products cannot be audited, so we audit the practice. We train the families
used in low-field SR on real paired data under standard recipes: Res-SRDiff \citep{safari2025ressrdiff}
(diffusion), DISGAN \citep{wang2023disgan}, medSynthesis \citep{nie2017medical,nie2018medical}, and SR-CycleGAN
adapted to paired training \citep{zheng2022srcyclegan} (GANs), and ViT-Fuser \citep{oved2025vitfuser} used as a
direct 64 mT to 3T mapping (transformer). The audit covers 24 runs with held-out predictions:
18 GAN, 1 transformer, and 5 diffusion
(1 multicontrast T2w and 4 single-contrast T1w). The direct fabrication
measurement uses the single-contrast T1w diffusion runs on the four held-out T1w subjects, sub-0048 and sub-0064
(Zenodo) and sub-HYPE21 and sub-HYPE22 (OpenNeuro), with 50 posterior samples per subject
under z-score normalization and 5 under WhiteStripe \citep{shinohara2014}. Their recipe, the T2w and NIST recipes,
and the trained runs outside the audit are listed in Appendix~\ref{app:reconstruction-details}.

\textbf{External arm.} The publicly released single-input SynthSR model \citep{iglesias2021synthsr} is audited
with identical measurements on the identical held-out subjects. It is not the dual-input low-field variant of
\citet{iglesias2023lowfield}, which needs a co-registered 64 mT T2w that these subjects lack.

\section{Results}
\label{sec:results}

\begin{table}[tbp]
\caption{Headline measurements (full table with controls: Table~\ref{tab:t1}). \HP{} is $1/(2k_c)$, \FP{} is
$1/k_c$; a dash means the row is tabulated in one convention only.}
\label{tab:main}
\centering\small
\resizebox{\linewidth}{!}{%
\begin{tabular}{@{}lllllll@{}}
\toprule
Quantity & Role & Dataset & $n$ & mm \HP{} & mm \FP{} & CI \HP{} \\
\midrule
Ceiling, cross-FSC half-bit & measurement & Zenodo T1w & 8 & 3.157 & 6.313 & 2.46--3.61 \\
Ceiling, cross-FSC half-bit & measurement & OpenNeuro T1w & 11 & 4.814 & 9.628 & 3.91--9.38 \\
Ceiling, cross-FSC half-bit & measurement & Zenodo T2w & 10 & 3.187 & 6.373 & 1.90--3.67 \\
Anisotropy, in / through plane & measurement & Zenodo T1w & 8 & -- & 6.13 / 11.07 & -- \\
\midrule
Gap vs S-KSP+N (primary) & measurement & Zenodo T1w & 8 & 1.162 & 2.324 & -- \\
Gap vs S-BIC+N / S-GAU+N & measurement & Zenodo T1w & 8 & 0.921 / 0.937 & 1.842 / 1.873 & -- \\
Gap vs S-KSP+N & measurement & OpenNeuro T1w & 11 & 3.115 & 6.231 & -- \\
\bottomrule
\end{tabular}}
\end{table}

\subsection{The Input Carries Subject Information Only Down to a Measured Ceiling}
\label{sec:ceiling}

\begin{figure}[tbp]
\centering
\begin{minipage}[t]{0.4\linewidth}\centering\figslot{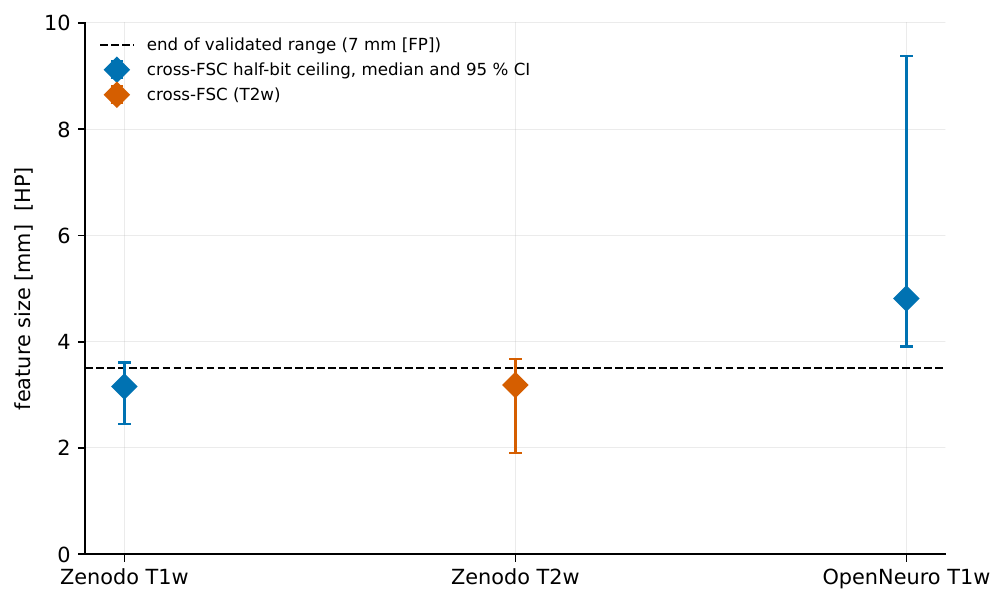}\end{minipage}\hfill
\begin{minipage}[t]{0.55\linewidth}\centering\figslot{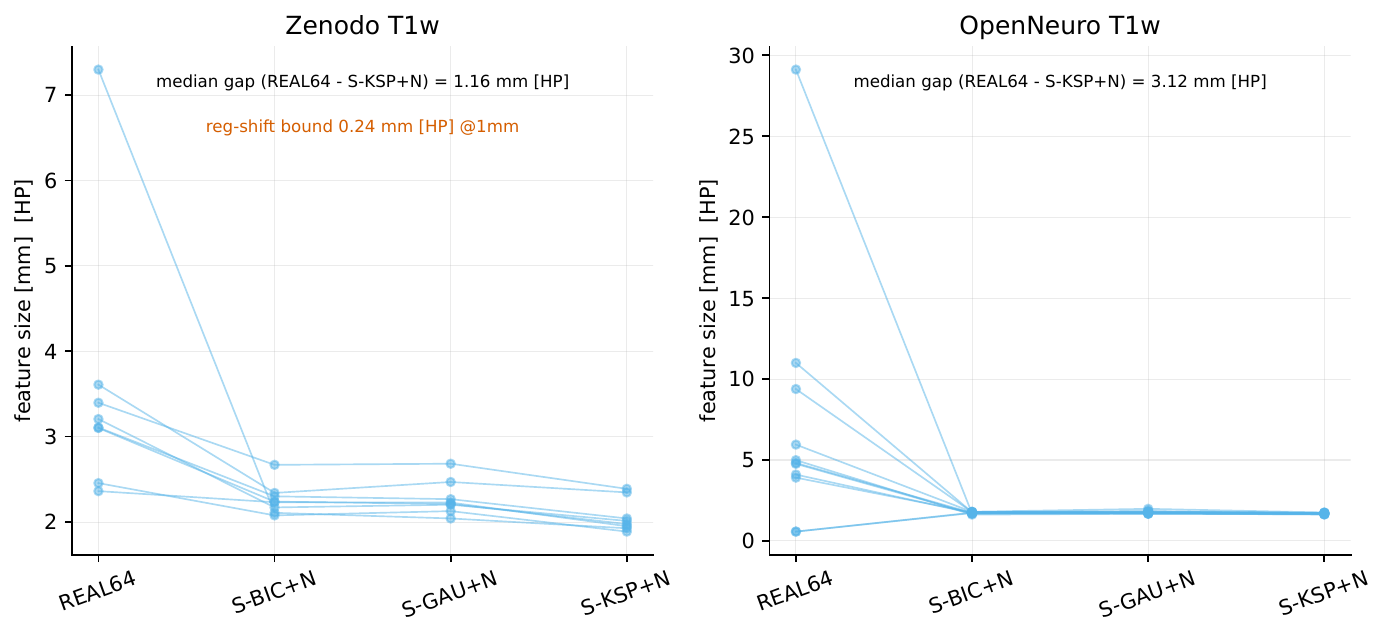}\end{minipage}
\caption{Left: \capFfour{} Right: \capFGAP}
\label{fig:f4}
\end{figure}

On Zenodo T1w the cross-FSC between the real 64 mT scan and the 3T scan of the same subject crosses the
half-bit threshold at 3.157~mm~\HP{} (6.313~mm~\FP{}, CI
2.46--3.61~mm~\HP{}, $n=8$; Table~\ref{tab:main}, Figure~\ref{fig:f4}, left; curves in Figure~\ref{fig:protocol}). In plain terms: judged over the
whole brain, the finest structure the two scans of one person still share is about 3~mm half-pitch; finer than
that, agreement falls below the half-bit threshold and approaches the band that subject-swapped pairs produce
(Figure~\ref{fig:protocol}). On OpenNeuro T1w the crossing is 4.814~mm~\HP{}
(9.628~mm~\FP{}, CI 3.91--9.38). On Zenodo T2w the crossing is 3.187~mm~\HP{},
close to T1w on the same scanner, so the ceiling belongs to the acquisition rather than the contrast.
The OpenNeuro T2w ceiling is registration-limited, qualitative (Mattes mutual information median around
$-0.45$, weak in a convention where better alignment is more negative, and beyond the FSC-validated
range), and we quote no millimetre value for it.

On NIST the cross-FSC is degenerate under EPI
distortion (Appendix~\ref{app:practice}); single-image Fourier ring correlation (1FRC, a resolution read from one image, \citealp{koho2019frc}) gives 5.06 (64 mT) and
3.63~mm~\FP{} (3T), a self-resolution and not a ceiling. The ceiling is anisotropic: the in-plane
sector crosses at 6.13~mm~\FP{} (3.07~mm~\HP{}) and the through-plane
sector at 11.07~mm~\FP{} (5.54~mm~\HP{}), reflecting the
$1.6\times1.6\times5.0$~mm acquisition; the headline 3.157~mm~\HP{} is the
3D shell average.

\textbf{Scope of the ceiling.} The ceiling is judged over the whole brain with the half-bit criterion. The band
readout of Section~\ref{sec:models} sets a lower bar on a smaller region: a scale counts as subject-specific if the
image matches its own subject's 3T better than a stranger's by more than 0.05. Applied to the raw 64 mT input, it
finds subject information somewhat finer than the whole-brain value on two subjects,
5.0~mm~\FP{} on sub-0048 (whole-brain anchor 6.9) and
6.0~mm~\FP{} on sub-HYPE21 (anchor 10.9), consistent with high-contrast
structure such as the ventricle margins. Finer than the ceiling, the input therefore carries little subject
information averaged over the brain but can still carry some at a few high-contrast edges; this is why every model
below is compared with its own raw input, read the same way, and not with the ceiling alone.
Figure~\ref{fig:finfo} (Appendix~\ref{app:figures}) shows the ceiling in image space.

\subsection{Synthetic Degradation Preserves What Real Acquisition Destroys}
\label{sec:synthetic}

Real 64 mT acquisition is coarser than the noise-matched k-space-truncation synthetic by
1.16~mm~\HP{} (2.32~mm~\FP{}) on Zenodo T1w, exceeding a registration
bound of 0.24~mm~\HP{} (0.47~mm~\FP{}), so synthetic degradation
pipelines preserve fine information that real low-field acquisition destroys (Figure~\ref{fig:f4}, right).
k-space truncation is the physically principled band-limit model, so its gap is primary; the softer S-BIC+N and
S-GAU+N give 0.921 and 0.937~mm~\HP{}, both above the bound, and on
OpenNeuro T1w the gap is 3.115~mm~\HP{} ($n=11$), in line with its coarser
ceiling. Two OpenNeuro subjects read at the grid cap (0.58~mm~\HP{}, the finest scale the analysis grid can express) despite good registration, a
physically implausible value for 64 mT; they are kept in the $n=11$ median at the capped
value rather than dropped after the fact; dropping them would move the median ceiling to 4.99~mm~\HP{}
(Figure~\ref{fig:f4}, right). Both OpenNeuro choices, excluding the null-crossing subject and keeping the capped
ones, move the ceiling finer and are therefore conservative for Claim 1.
Because synthetics inherit 3T contrast and the real 64 mT scan does not, part of the gap could in principle come
from the nonlinear way tissue contrast changes between field strengths rather than from lost information. Three
observations argue against this being the driver. The crossing's spread across intensity normalizations is below
the gap (D1 ratio 0.79), and the gap is positive under every normalization we ran
(1.16, 1.46 and 1.83~mm~\HP{} on Zenodo, 3.11, 3.11 and 5.11 on OpenNeuro),
the tabulated value being the smallest on Zenodo and within 0.01~mm of the smallest on
OpenNeuro. T1w and T2w, whose contrast mappings to 3T differ, give the same ceiling on the same scanner
(6.313 versus 6.373~mm~\FP{}). Giving the synthetic low-field contrast, by
matching each tissue class of the 3T to the 64 mT scan's median and spread before truncation, narrows the gap
only to 0.91~mm~\HP{}, positive on all 8 subjects (minimum
0.27) and above the registration bound. A model trained and tested on synthetic
pairs therefore sees subject structure in a band that a real 64 mT scan leaves empty; recovering it is recovery
on the benchmark and invention in deployment, and the benchmark cannot tell which.

\subsection{Trained Models Fabricate at Fine Scales, and Trust Tools Do Not Flag It}
\label{sec:models}

This section asks whether trained models respect the ceiling, and whether the tools the field trusts, metrics
and uncertainty maps, would notice if they did not.
The claim rests on three layers, strongest first. (1) \emph{Input-side bound}: the ceiling is a property of
the input, measured directly; no reconstruction can carry more subject information than its input, so
subject-specific detail finer than the ceiling can only be inferred from coarser structure through the prior.
(2) \emph{Direct demonstration} on the diffusion arm ($n=4$ held-out subjects,
2 of them certified by every check).
(3) \emph{Coverage}: 24 trained runs of five architectures, on which faithfulness is
computable for the 4 single-contrast diffusion runs and not computable on the remaining
20 for documented reasons.
In short: the input carries no subject information at fine scales, the stranger test shows that these models do
not infer it either, and their metrics and uncertainty maps do not register the difference.

\begin{figure}[tbp]
\centering
\figslot[0.84\linewidth]{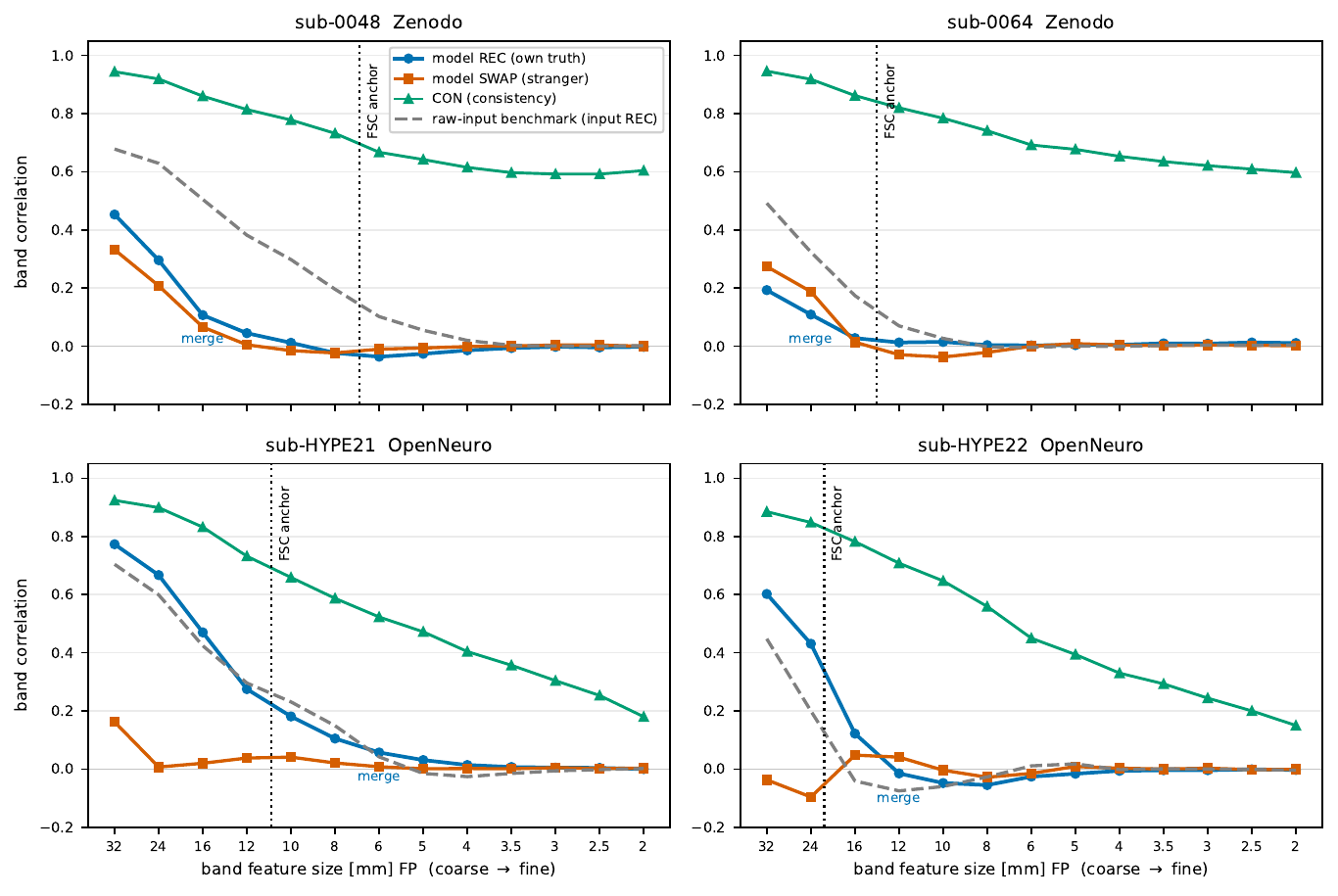}
\caption{\textbf{Fabrication against the raw-input benchmark.} REC (output vs own 3T), SWAP (output vs
stranger 3T) and CON (between samples) by band feature size \FP{} for the four held-out subjects (z-score arm,
50 samples); grey dashed: raw-input benchmark; dotted: whole-brain FSC anchor. Merges in Table~\ref{tab:ladder}.}
\label{fig:fp4}
\end{figure}

\textbf{Direct demonstration.} The T1w arms carry the direct measurement because their held-out inputs have
usable per-subject anchors, and every model is read against its own subject's anchor and raw-input benchmark.
All four subjects pass G-M1 and G-M4. G-M2 passes sub-0048 and sub-HYPE21 and fails sub-0064 and sub-HYPE22,
whose anchors (14.04 and 21.6~mm~\FP{}) lie outside the dataset
intervals; on OpenNeuro this gate is frame-dependent, so the per-subject anchor on the validated frame is the
operative check. G-M3 passes every subject except sub-0064, where the stranger's 3T correlates with the input to
about the same scale as the subject's own, so on that subject the readout cannot separate fabrication from
misalignment of its own 3T. Two subjects, one per dataset, are therefore certified; the other two are shown for
completeness (Table~\ref{tab:ladder}, Appendix~\ref{app:gates}), and the finding below is the same on all four. On
T2w the held-out subjects fail G-M2 (Zenodo) or too few pass (OpenNeuro, one of four), so all
13 T2w runs are not computable.
REC merges with SWAP at 16.0,
24.0, 6.0 (posterior mean 5.0) and
12.0~mm~\FP{} for sub-0048, sub-0064, sub-HYPE21 and sub-HYPE22, against raw-input
benchmarks of 5.0, 12.0, 6.0 and
16.0~mm~\FP{} (Table~\ref{tab:ladder}, Figure~\ref{fig:fp4}). Finer than each subject's
merge scale the output is no more correlated with the subject's own anatomy than with a stranger's.
At fine scales all four subjects show REC at SWAP: fabrication. On Zenodo the fabricated structure is internally
consistent across samples (CON 0.59 to 0.95), confident fabrication; on OpenNeuro inter-sample
consistency itself decays at fine bands (CON near 0.15), diffuse fabrication. Neither form is
distinguishable from recovery by any metric or uncertainty map we audit.

\textbf{Exceedances and the two datasets.} Two merges in the diffusion arm fall one band finer than their
raw-input benchmark (both single-band, both within 0.025 of threshold, both on the
coarse-recovery subjects); they are the coarse-scale denoising the benchmark allows (Section~\ref{sec:protocol}),
not information creation. No model shows sustained subject-specific agreement at scales where the input carries
none; the one further merge finer than a benchmark is a negative-SWAP artifact with REC at zero (SynthSR, below).
On Zenodo the models fabricate at scales where the raw input still carries subject information, and for sub-0048 the output matches the truth less
than the raw input at every scale from 32 to 8 mm: the model degrades structure the raw input still resolves,
failing even the denoising role the OpenNeuro arm demonstrates. On OpenNeuro the models recover coarse
ventricle-scale structure to the input's limit and fabricate finer than it; this recovery also serves as a positive
control that the measurement detects genuine recovery when present.

\textbf{Controls.} The S-KSP+N oracle recovers strongly (merge 2.5~mm~\FP{} on OpenNeuro),
so the readout reports recovery when it is present; on Zenodo it never merges within the grid, because the
truncation is a box in k-space and the bands are an octave wide, so the oracle keeps a band correlation of
0.70 at 4~mm and 0.40 at 2.5~mm although its own half-bit crossing is
2.7~mm~\FP{}. Under WhiteStripe (5 seeds) merges are
10 / 24 / 6 / 16~mm~\FP{}, the same picture. Halving or doubling the 0.05 threshold moves
each merge by at most one band, except the sub-0048 diffusion merge at the lower threshold (16 to 10~mm), and at
every threshold the models merge coarser than their raw input on Zenodo and at or within one band of it on
OpenNeuro. Misregistration cannot produce the merges: shifting the true 3T by 1~mm in plane or by one slice keeps
its band correlation with itself at 0.70 to 0.99 across the 6 to 24~mm bands where the
models merge, against 0.03 to 0.12 for the Zenodo models and
0.06 to 0.47 for the OpenNeuro models; at 3~mm and finer a 1~mm shift alone lowers it to
0.4 or below, so the finding rests on the merge scales, not on the finest bands. Replacing
the frozen stranger by each same-dataset training subject (8 on Zenodo,
21 on OpenNeuro, each rigidly registered into the held-out frame) gives diffusion merges of
8 to 16 / 16 to 24 / 4 to 10 / 10 to 16 and raw-input merges of
4 to 5 / 10 to 24 / 5 to 8 / 8 to 24~mm~\FP{}, each interval containing the frozen-pair
merge; these strangers were seen by the model in training, so the pool is a robustness check. Resampling the
Zenodo models with their 15-step training chain leaves both z-score merges unchanged
(16 and 24~mm~\FP{}) and their uncertainty controls within 0.01 of the
values below, and moves the WhiteStripe merges to 12 and 24~mm~\FP{}, no finer than before.

\begin{table}[tbp]
\caption{Control ladder for the direct measurement, all mm~\FP{}. Anchor: whole-brain cross-FSC of the
subject's input. Raw-input benchmark: the registered 64 mT input through the identical band readout (a
correlation benchmark, not an information bound). Oracle: S-KSP+N of the subject's own 3T (``--'': no merge
within the grid). Diffusion: z-score arm, 50 samples (posterior mean in parentheses where it
differs). SynthSR: deterministic. G-M2, G-M3: audit status of the input (Section~\ref{sec:protocol},
Appendix~\ref{app:gates}); a subject passing both is certified.$^{\dagger}$}
\label{tab:ladder}
\small
\begin{tabular}{@{}llccccccc@{}}
\toprule
Subject & Dataset & G-M2 & G-M3 & Anchor & Raw-input & Oracle & Diffusion & SynthSR \\
\midrule
sub-0048 & Zenodo & pass & pass & 6.9 & 5.0 & -- & 16.0 & 12.0 \\
sub-0064 & Zenodo & fail & marginal & 14.04 & 12.0 & -- & 24.0 & 24.0 \\
sub-HYPE21 & OpenNeuro & pass & pass & 10.9 & 6.0 & 2.5 & 6.0 (5.0) & 6.0 \\
sub-HYPE22 & OpenNeuro & fail & pass & 21.6 & 16.0 & 2.5 & 12.0 & 10.0 \\
\bottomrule
\end{tabular}
\par\vspace{2pt}
{\footnotesize\raggedright $^{\dagger}$Two diffusion merges fall one band finer than the benchmark (sub-HYPE22
12.0 vs 16.0; sub-HYPE21 posterior mean 5.0
vs 6.0), each decided by one band clearing the 0.05 threshold by
0.023 and 0.016: coarse-scale denoising, not information creation.
SWAP uses one frozen stranger per subject (sub-0048 with sub-0064, sub-HYPE21 with sub-HYPE22). G-M2 on OpenNeuro
is frame-dependent and G-M3 on sub-0064 passes by 0.01~mm under the ladder anchor
recipe and fails by 5.0~mm under the gate recipe (Appendix~\ref{app:gates}). All arms are read
on one ROI per subject: the 3T foreground mask intersected with the model's coverage and the sampled slices
(Appendix~\ref{app:reconstruction-details}).
Under a SWAP clipped at zero ($\mathrm{REC}-\max(\mathrm{SWAP},0)$) every merge in this table is unchanged except the
SynthSR sub-HYPE22 entry, which becomes 16.0, equal to its benchmark, and the sub-0048
raw-input benchmark, which becomes 4.0 because its 5.0~mm band sits
exactly at the 0.05 threshold. A merge of 24.0 is the coarsest the rule can
report and means no subject-specific agreement at any reportable band.}
\end{table}

\textbf{External model.} On the same subjects and readout, SynthSR merges at
12 / 24~mm~\FP{} on Zenodo, coarser than the raw-input benchmarks of
5.0 / 12.0, and at 6.0 /
10.0~mm~\FP{} on OpenNeuro against benchmarks of 6.0 /
16.0; the one merge finer than its benchmark comes from a negative SWAP with REC at zero and
moves to the benchmark, 16.0~mm~\FP{}, when SWAP is clipped at zero
(Appendix~\ref{app:synthsr}). Its apparent sharpness, 5.2 to 6.1~mm~\FP{}, sits at the
input ceiling rather than beyond it, so it invents less fine detail than the trained runs while, on Zenodo, still
failing to match the subject where the input would allow it (on sub-0064 its output aligns poorly with the 3T, so
that merge is read with care). The behavior is therefore not a property of our implementations.

\textbf{Uncertainty.}
Sample-variance uncertainty flags reconstruction difficulty, not fabrication. The variance map carries real
spatial error signal: the voxel AUROC of variance flagging top-decile band error is
0.771 on bands finer than 4~mm~\FP{} and 0.677 on
bands of 4~mm~\FP{} and coarser (chance level 0.5), and its Spearman correlation with
error is 0.390 and 0.293, surviving control for band energy at partial
0.260 and 0.205. It is, however, largely non-specific: in every
band of the original registration frame its correlation with a stranger's error is at least
two thirds of its correlation with the subject's own (RHO\_own 0.390 versus RHO\_stranger 0.307 on the fine
bands; Table~\ref{tab:t3}), so it locates hard regions rather than fabricated content (Figure~\ref{fig:fsharp},
left).
Recomputed on the validated frames and the common ROI, per subject, the
fine-band AUROC is 0.69 to 0.79 and the correlation with error 0.30 to 0.39, and
a known-fabricated output, the frozen stranger's own posterior samples read against this subject's 3T, receives
the same variance-error agreement as the subject's own output on Zenodo (AUROC 0.71 and 0.72
against 0.71 and 0.69) and less on OpenNeuro (0.59 and 0.41 against
0.79 and 0.78); the conclusion is unchanged.

\textbf{Coverage.} The audit spans 24 trained runs of five architectures (Table~\ref{tab:t2}).
Faithfulness is computable on the four single-contrast T1w diffusion runs and fails on all four; the other
20 carry a reason code (13 T2w runs, NC-ANCHOR or NC-XDEV; 7
NIST runs, NC-EPI). Every trained run renders detail finer than its input ceiling (apparent sharpness
0.97--3.77~mm~\FP{}, Figure~\ref{fig:fsharp}, right), whether or not that detail can be checked.
Because faithfulness fails wherever it can be computed, the audit cannot test whether PSNR or SSIM track it across
models, but on the four direct-arm subjects the metrics barely separate fabricating outputs from the unfabricated
input: PSNR puts the diffusion posterior mean within 1.4~dB of the raw input, against an oracle
lead of 4.7 to 7.7~dB, and LPIPS scores it as good as or better than the raw input on every
subject, including both Zenodo subjects, where its merge is far coarser.

\section{Limitations and Lessons}
\label{sec:limitations}

\textbf{Limitations.} The direct demonstration rests on $n=4$ held-out subjects, every held-out
subject with a trained single-contrast diffusion model and ground truth; we show per-subject curves and rely on
agreement across datasets, normalizations, strangers and thresholds instead of a pooled test, and the WhiteStripe
rows use 5 seeds. 
Only 2 of the four, one per dataset, are certified by every audit check; the result is the same
on all four.
The merge is read against one
frozen stranger per subject, and the stranger pool that brackets it uses training subjects the model has seen. The
diffusion checkpoints were selected on validation PSNR of the same held-out subjects
(Appendix~\ref{app:reconstruction-details}), which favours agreement with their own 3T and so works against the
fabrication finding, though it may contribute to the two diffusion exceedances. The reported samples use 4 steps while
the Zenodo models were trained with a 15-step chain; 
resampling with 15 steps leaves the
Zenodo merges unchanged or coarser.
Faithfulness is not computable on 20 of
24 runs, further trained runs with held-out predictions are outside the audit
(Appendix~\ref{app:reconstruction-details}), proprietary products remain unauditable, and the public SynthSR we
audit is the single-input model, not the dual-input low-field variant. A different scanner or sequence will have a
different ceiling, which is why we contribute a protocol and not a constant; an accurate forward model of a
specific scanner, learned or simulated, would let the ceiling be derived analytically and models be trained
against a known operator. The
strongest layer, the input ceiling, depends on no model and on none of the four subjects.

\textbf{Lessons for practice} (Appendix~\ref{app:practice}). Each documented failure implies a check: report a volumetric resolution only beside a grid-invariance check; report a cross-field ceiling as not computable when rigid registration cannot align the pair; check any fixed test split against the input anchor (G-M2) before reporting metrics on it; validate the registration of every held-out subject before reading a model against its input; and gate a
held-out subject only on a frame on which both its anchor and the dataset ceiling are measured, since a
dataset-level gate is undefined across frames.

\section{Conclusion}
\label{sec:conclusion}
A real 64 mT scan limits what any model can recover about the subject, and the limit can be measured: about 3 to
4~mm half-pitch in plane here, coarser than the detail synthetic benchmarks preserve and coarser than the detail
current models render. Recoverability claims for low-field super-resolution should therefore be tested against the
measured input ceiling and a stranger control rather than PSNR, SSIM, or sample variance alone. We will release the protocol code so that both checks can be run on any model. Its two tools, cross-modality FSC
and the stranger-controlled band readout, apply to any setting with paired acquisitions.

\section*{Acknowledgments} This work used computing resources from the National Research Platform (NRP) Nautilus
cluster and the NCSA Delta system. We thank the NRP and NCSA teams for their support.

\section*{Reproducibility Statement}
All datasets are public (Section~\ref{sec:data}). The protocol, gate definitions (Table~\ref{tab:gates}),
algorithms (Appendix~\ref{app:alg}), frozen splits, stranger lists, seeds, sampler settings, result files and a
verification script that regenerates every table will be made available online with the code; the external
SynthSR run is documented in Appendix~\ref{app:synthsr}.

\section*{Ethics Statement}
All data are public, de-identified human MRI released by their providers for research (Section~\ref{sec:data});
no new data were collected. The work evaluates existing methods and argues for caution in their clinical use; it
introduces no new generative model.

\section*{Use of Large Language Models}
Large language model assistants were used to help write and debug analysis and plotting code, to check the
consistency of numbers between result files, tables, and text, and to edit the prose of this paper. All
experiments, results, and claims were designed, verified, and approved by the authors, who take full
responsibility for the content.

\bibliography{trimmed_refs}
\bibliographystyle{iclr2027_conference}

\appendix

\section{Algorithms}
\label{app:alg}

\begin{algorithm}[h]
\caption{Gated 3D cross-FSC with half-bit and empirical-null readout}
\label{alg:fsc}
\begin{algorithmic}[1]
\Require registered volumes $X,Y$ on one grid (soft dilated brain mask $\times$ Hann window), voxel size $v$,
shell width $\Delta k=1/(Nv)$
\State $F_X\gets\mathrm{FFT}(X)$, $F_Y\gets\mathrm{FFT}(Y)$; radial frequency grid $K$ in cycles/mm
\For{each shell $S_j=\{K\in[j\Delta k,(j+1)\Delta k)\}$}
  \State $\mathrm{FSC}_j\gets \mathrm{Re}\sum_{S_j}F_X\overline{F_Y}\,/\,\sqrt{\sum_{S_j}|F_X|^2\sum_{S_j}|F_Y|^2}$;\ \ $n_j\gets|S_j|$
  \State $T_j\gets(0.2071+1.9102/\sqrt{n_j})/(1.2071+0.9102/\sqrt{n_j})$ \Comment{half-bit threshold \citep{vanheel2005halfbit}}
\EndFor
\State $k_c\gets$ interpolated frequency where FSC falls below $T$ and stays below for two consecutive shells
\State repeat against the empirical null curve from subject-swapped pairs
\State \Return $1/k_c$ \FP{}, $1/(2k_c)$ \HP{}; anisotropic variant uses in-plane and through-plane sectors
\end{algorithmic}
\end{algorithm}

\begin{algorithm}[h]
\caption{Decorrelation analysis (single-image self-resolution, port of \citet{descloux2019decorr})}
\label{alg:decorr}
\begin{algorithmic}[1]
\Require image $I$; $N_r$ radii; $N_g$ high-pass strengths
\State remove DC; $I_n\gets\mathrm{FFT}(I)/|\mathrm{FFT}(I)|$
\For{each Gaussian high-pass strength $g$}
  \For{each radius $r\in[0,\text{Nyquist}]$ with low-pass mask $M_r$}
    \State $d_{g}(r)\gets\mathrm{Re}\sum \mathrm{FFT}(I_g)\overline{I_n}M_r/\sqrt{\sum|\mathrm{FFT}(I_g)M_r|^2\sum|I_nM_r|^2}$
  \EndFor
  \State $r_g\gets$ highest-frequency local maximum of $d_g$
\EndFor
\State \Return $1/\max_g r_g$ \FP{}
\end{algorithmic}
\end{algorithm}

\begin{algorithm}[h]
\caption{Band-resolved faithfulness: REC, SWAP, CON, merge scale, and uncertainty}
\label{alg:rec}
\begin{algorithmic}[1]
\Require samples $y_1,\dots,y_N$ for a held-out subject; own 3T $t$; stranger 3T $t'$ (frozen pair list); bands
$s_1>\dots>s_B$ (mm \FP{}); $\mathrm{bp}_s$: one-octave annulus band-pass centred at $s$, applied per 2D slice;
ROI: 3T foreground $\cap$ model coverage $\cap$ sampled slices; $\mathrm{corr}$ is the Pearson correlation over
ROI voxels
\For{each band $s$}
  \State $B_t\gets\mathrm{bp}_s(t)$, $B_{t'}\gets\mathrm{bp}_s(t')$, $B_n\gets\mathrm{bp}_s(y_n)$
  \State $\mathrm{REC}(s)\gets\mathrm{mean}_n\,\mathrm{corr}(B_n,B_t)$;\ \ $\mathrm{SWAP}(s)\gets\mathrm{mean}_n\,\mathrm{corr}(B_n,B_{t'})$
  \State $\mathrm{CON}(s)\gets\mathrm{mean}_{n<m}\,\mathrm{corr}(B_n,B_m)$
  \State $\mathrm{VAR}\gets\mathrm{var}_n(y_n)$ per voxel;\ \ $\mathrm{ERR}\gets|\mathrm{bp}_s(\bar y)-B_t|$;\ \ RHO $\gets$ Spearman(VAR, ERR)
\EndFor
\State merge $\gets$ coarsest $s$ below $s_1$ such that $\mathrm{REC}-\mathrm{SWAP}\le0.05$ at $s$ and at the next
finer band \Comment{clipped variant: $\mathrm{REC}-\max(\mathrm{SWAP},0)$}
\State the same readout is applied to the registered 64 mT input (raw-input benchmark), to S-KSP+N (oracle), and to
SynthSR; for a single deterministic image $N=1$ and CON is not defined
\end{algorithmic}
\end{algorithm}

\FloatBarrier
\section{Validation Gates}
\label{app:gates}
Table~\ref{tab:gates} lists the known-answer gates, their pre-specified criteria, and their outcomes, including
the two criteria that are not met.
\begin{table}[H]
\caption{Known-answer gates run on Zenodo before any reported value (3 subjects). Pass criteria are
pre-specified; decision D1 is reported in Section~\ref{sec:protocol}.}
\label{tab:gates}
\centering\small
\resizebox{\linewidth}{!}{\begin{tabular}{@{}llll@{}}
\toprule
Gate & Test & Pre-specified criterion & Outcome \\
\midrule
A1 & recovery of an imposed cutoff (3, 4, 5, 6, 8 mm [FP]) & within 10\% & FSC: 2.5--7.5\% error at 3--6 mm (pass over the operative range); 9--11\% at 8 mm (fails the strict \\
   &  &  & criterion; brain field of view); decorrelation fails (0 of 15) and is kept qualitative \\
A2 & 3T self-split (binomial 1FRC) vs native 2.0 mm & within 25\% & 2.44--2.77 mm (22--38\%): fails; used only as a fine upper bound, far finer than the 6.31 mm [FP] crossing \\
A3 & grid invariance (1.0 vs 1.25 mm grid) & within 10\% & within 3\%: pass \\
A4 & registration robustness (0 to 2 mm injected shift) & bound below every gap & crossings move 0.02--0.57 mm [FP]; a 1 mm shift moves the crossing 0.47 mm [FP], below the 2.32 mm [FP] gap \\
A5 & response to added blur (FWHM 2--6 mm) & strictly monotone & FSC strictly monotone: pass; decorrelation fails \\
-- & cross-check against an independent FRC implementation & agreement & 3.79 vs 3.58 mm on a 4.0 mm imposed cutoff (5.7\%) \\
\bottomrule
\end{tabular}
}
\end{table}

Table~\ref{tab:auditgates} gives the four audit checks of Section~\ref{sec:protocol} for the held-out subjects of
the direct arm, and Table~\ref{tab:frames} the OpenNeuro T1w gate under each frame pairing. The anchors of
Table~\ref{tab:ladder} use the ladder recipe; the gate recipe (whole-brain FSC with z-score intensity normalization) differs by
0.5 to 1.1~mm and gives the same pass or fail on every subject.
\begin{table}[H]
\caption{Audit checks on the direct-arm inputs, all mm~\FP{}. G-M1: 3T self-split fine bound (1FRC on Zenodo,
checkerboard split on OpenNeuro). G-M2: anchor inside the dataset CI (Zenodo T1w 4.91--7.22,
OpenNeuro T1w 7.82--18.77, both on the E1 frame). G-M3: input cross-FSC against the frozen
stranger's 3T, required at least one band coarser than the own anchor or absent, under the ladder and the gate
anchor recipes. G-M4: deterministic rerun. Certified: G-M2 and G-M3 both pass.}
\label{tab:auditgates}
\centering\small
\resizebox{\linewidth}{!}{%
\begin{tabular}{@{}lccccccc@{}}
\toprule
Subject & G-M1 & Anchor (ladder / gate) & G-M2 & Anchor $\le$ 7 & G-M3 required & G-M3 (ladder / gate) & Certified \\
\midrule
sub-0048 & 2.53 & 6.90 / 6.41 & pass & yes & $\ge$ 8 & 17.6 / 36.4 pass & yes \\
sub-0064 & 2.42 & 14.04 / 14.60 & fail & no & $\ge$ 16 & 16.01 / 10.97 marginal & no \\
sub-HYPE21 & 1.16 & 10.90 / 11.52 & pass & no & $\ge$ 12 & absent / 69.3 pass & yes \\
sub-HYPE22 & 1.16 & 21.60 / 22.72 & fail & no & $\ge$ 24 & 47.8 / 45.0 pass & no \\
\bottomrule
\end{tabular}}
\end{table}
\begin{table}[H]
\caption{OpenNeuro T1w gate G-M2 under each frame pairing, mm~\FP{}. The paper's evaluation mixes frames (CI
on E1, anchors on the validated v3 frame). Like against like on E1 the held-out inputs do not register; like
against like on v3 the dataset ceiling is degenerate (12 subjects, foreground mask). No
pairing passes both subjects.}
\label{tab:frames}
\centering\small
\begin{tabular}{@{}llccc@{}}
\toprule
Frames (CI / anchors) & Dataset ceiling and CI & sub-HYPE21 anchor & sub-HYPE22 anchor & Outcome \\
\midrule
E1 / v3 (paper) & 9.628 (7.82--18.77) & 10.90 & 21.60 & pass / fail \\
E1 / E1 & 9.628 (7.82--18.77) & 51.4 (no crossing, gate recipe) & 55.8 / 101.3 & fail / fail \\
v3 / v3 & 158 (22--512) & 10.90 & 21.60 & fail / pass \\
\bottomrule
\end{tabular}
\end{table}

Table~\ref{tab:knownanswer} calibrates the merge readout of Algorithm~\ref{alg:rec} on a known answer. For each
held-out subject $A$ with frozen stranger $B$, $F_c=\mathrm{low}_c(A)+\mathrm{high}_c(B)$ with complementary hard
filters at cutoff $c$ (the two parts sum back to the full image), read on the common ROI with the ladder rule;
the recovered merge should equal $c$. A noisy repeat (Rician noise at the input's signal-to-noise ratio) gives
identical merges in every case, and clipped and unclipped merges agree in every case. The band that straddles
the cutoff mixes both subjects' content, and whether it merges is decided by which subject's 3T carries more
energy there, so the error is always toward the finer side and never coarser.
\begin{table}[H]
\caption{Known-answer test of the merge readout, recovered merge in mm~\FP{} (error in grid steps). End cases:
$B$ alone merges at 24~mm, the coarsest reportable band; $A$ alone never merges.}
\label{tab:knownanswer}
\centering\small
\begin{tabular}{@{}lccccccc@{}}
\toprule
Subject & $c=16$ & $c=12$ & $c=8$ & $c=6$ & $c=4$ & $B$ alone & $A$ alone \\
\midrule
sub-0048 & 16 (0) & 12 (0) & 8 (0) & 6 (0) & 4 (0) & 24 & none \\
sub-0064 & 12 (1) & 10 (1) & 6 (1) & 5 (1) & 3 (2) & 24 & none \\
sub-HYPE21 & 12 (1) & 10 (1) & 6 (1) & 5 (1) & 3 (2) & 24 & none \\
sub-HYPE22 & 16 (0) & 12 (0) & 8 (0) & 6 (0) & 4 (0) & 24 & none \\
\bottomrule
\end{tabular}
\end{table}

\FloatBarrier
\section{Full Tables}
\label{app:tables}
Table~\ref{tab:t1} gives every measurement with its controls and source, Table~\ref{tab:t2} the per-run audit,
and Table~\ref{tab:t3} the uncertainty controls per band.

\begin{table}[H]
\caption{\capTthree}
\label{tab:t3}
\centering\small
\begin{tabular}{lllll}
\toprule
band centre mm [FP] or group & RHO\_own & partial RHO & RHO\_stranger & AUROC \\
\midrule
8.0 & 0.171 & 0.123 & 0.115 & 0.566 \\
6.0 & 0.270 & 0.189 & 0.219 & 0.659 \\
5.0 & 0.342 & 0.239 & 0.280 & 0.725 \\
4.0 & 0.388 & 0.269 & 0.315 & 0.758 \\
3.5 & 0.395 & 0.271 & 0.318 & 0.763 \\
3.0 & 0.392 & 0.265 & 0.313 & 0.766 \\
2.5 & 0.387 & 0.258 & 0.304 & 0.773 \\
2.0 & 0.387 & 0.245 & 0.294 & 0.781 \\
\midrule
coarse group: mean of 8.0--4.0 & 0.293 & 0.205 & 0.232 & 0.677 \\
fine group: mean of 3.5--2.0 & 0.390 & 0.260 & 0.307 & 0.771 \\
\bottomrule
\end{tabular}

\end{table}

\begin{sidewaystable}
\caption{\capTone}
\label{tab:t1}
\centering\tiny
\resizebox{\linewidth}{!}{

\begin{tabular}{llllllll}
\toprule
claim & quantity (role) & dataset & n & mm [HP] & mm [FP] & CI [HP] & controls / readout \\
\midrule
1 & ceiling: cross-FSC half-bit (measurement) & Zenodo T1w & 8 & 3.157 & 6.313 & 2.46-3.61 & A1 (3--6 mm), A3--A5; A2 fine bound only; null, FRC; half-bit crossing \\
1 & decorrelation, 64 mT alone (consistency check) & Zenodo T1w & 8 & 4.172 & - & - & qualitative; highest-frequency local max \\
1 & blur-sweep $k_c$, 3T alone (consistency check) & Zenodo T1w & 8 & 4.168 & - & - & consistency check; blur-sweep half-drop \\
1 & ceiling: cross-FSC (measurement) & Zenodo T2w & 10 & 3.187 & 6.373 & 1.90-3.67 & A1, null, D1; half-bit crossing \\
1 & ceiling: cross-FSC (measurement) & OpenNeuro T1w & 11 & 4.814 & 9.628 & 3.91-9.38 & A1 (3--6 mm), A3--A5; A2 fine bound only; beyond validated range ($>$7 mm [FP]); half-bit crossing \\
1 & ceiling: cross-FSC, registration-limited, qualitative & OpenNeuro T2w & 23 & - & - & - & A1; MI median -0.45; beyond validated range \\
1 & anisotropy in-plane / through-plane & Zenodo T1w & 8 & - & 6.13 / 11.07 & - & in-plane 2D vs through-plane 1D \\
2 & gap vs S-KSP+N (primary) & Zenodo T1w & 8 & 1.162 & 2.324 & - & exceeds registration bound 0.473 [FP]; difference of medians \\
2 & gap vs S-BIC+N & Zenodo T1w & 8 & 0.921 & 1.842 & - & noise-matched; difference of medians \\
2 & gap vs S-GAU+N & Zenodo T1w & 8 & 0.937 & 1.873 & - & noise-matched; difference of medians \\
2 & gap vs S-KSP+N & OpenNeuro T1w & 11 & 3.115 & 6.231 & - & noise-matched; difference of medians \\
neg & self-1FRC 64 mT / 3T (cross-FSC degenerate) & NIST & 19/20 & - & 5.06 / 3.63 & - & distortion-robust single-image readout \\
3 & merge, diffusion z-score (50 samples) & sub-0048 / 0064 / HYPE21 / HYPE22 & 4 & - & 16.0 / 24.0 / 6.0 (5.0) / 12.0 & - & two-band persistence; posterior mean in parentheses \\
3 & merge, diffusion z-score, 15-step sampling (50 samples) & sub-0048 / 0064 & 2 & - & 16.0 / 24.0 & - & training chain length; unchanged from 4 steps \\
3 & merge, diffusion WhiteStripe (5 samples) & sub-0048 / 0064 / HYPE21 / HYPE22 & 4 & - & 10.0 / 24.0 / 6.0 / 16.0 & - & two-band persistence \\
3 & merge, SynthSR (deterministic) & sub-0048 / 0064 / HYPE21 / HYPE22 & 4 & - & 12.0 / 24.0 / 6.0 / 10.0 & - & two-band persistence; HYPE22 exceedance from negative SWAP \\
3-ctrl & raw-input benchmark merge & sub-0048 / 0064 / HYPE21 / HYPE22 & 4 & - & 5.0 / 12.0 / 6.0 / 16.0 & - & correlation benchmark, not an information bound \\
3-ctrl & whole-brain FSC anchor & sub-0048 / 0064 / HYPE21 / HYPE22 & 4 & - & 6.9 / 14.04 / 10.9 / 21.6 & - & input cross-FSC per subject \\
3-ctrl & S-KSP+N oracle merge & HYPE21 / HYPE22 (Zenodo: none in grid) & 2 & - & 2.5 / 2.5 & - & readout detects recovery when present \\
3 & CON, inter-sample consistency (not mm) & Zenodo / OpenNeuro T1w & 2 / 2 & - & - & - & 0.59 to 0.95 (confident) / near 0.15 (diffuse) \\
3 & UQ AUROC finer than / at least 4 mm [FP] (not mm) & Zenodo + OpenNeuro T1w & 4 & - & - & - & original frame 0.77 / 0.68; validated frame, fine bands 0.69 to 0.79 per subject; partial Spearman, stranger control \\
\bottomrule
\end{tabular}}
\end{sidewaystable}

\begin{sidewaystable}
\caption{\capTtwo}
\label{tab:t2}
\centering\tiny
\resizebox{\linewidth}{!}{\begin{tabular}{lllllllll}
\toprule
run\_id & family & dataset & training pairing & apparent sharpness mm [FP] & PSNR & SSIM & LPIPS & faithfulness (merge mm [FP]) or reason code \\
\midrule
disgan\_zenodo\_minmax\_lpips\_wavelet & GAN & Zenodo T2w & 64mT$\to$3T T2w & 2.06 & 14.45 & 0.921 &  & NC-ANCHOR \\
disgan\_zenodo\_minmax\_balanced\_image & GAN & Zenodo T2w & 64mT$\to$3T T2w & 2.35 & 14.66 & 0.925 &  & NC-ANCHOR \\
medsynthesis\_zenodo\_minmax\_lpips\_wavelet & GAN & Zenodo T2w & 64mT$\to$3T T2w & 2.26 & 14.15 & 0.920 &  & NC-ANCHOR \\
medsynthesis\_zenodo\_minmax\_balanced\_image & GAN & Zenodo T2w & 64mT$\to$3T T2w & 3.77 & 13.49 & 0.918 &  & NC-ANCHOR \\
srcyclegan\_zenodo\_minmax\_lpips\_wavelet & GAN & Zenodo T2w & 64mT$\to$3T T2w & 2.06 & 13.65 & 0.912 &  & NC-ANCHOR \\
srcyclegan\_zenodo\_minmax\_balanced\_image & GAN & Zenodo T2w & 64mT$\to$3T T2w & 2.06 & 13.56 & 0.912 &  & NC-ANCHOR \\
disgan\_nist\_minmax\_lpips\_wavelet & GAN & NIST & 64mT$\to$3T DWI b0 & 1.60 & 28.41 & 0.931 & 0.043 & NC-EPI \\
disgan\_nist\_whitestripe\_balanced\_image & GAN & NIST & 64mT$\to$3T DWI b0 & 1.66 & 30.16 & 0.892 & 0.068 & NC-EPI \\
medsynthesis\_nist\_minmax\_lpips\_wavelet & GAN & NIST & 64mT$\to$3T DWI b0 & 0.97 & 30.29 & 0.944 & 0.030 & NC-EPI \\
medsynthesis\_nist\_whitestripe\_balanced\_image & GAN & NIST & 64mT$\to$3T DWI b0 & 2.02 & 32.27 & 0.904 & 0.057 & NC-EPI \\
srcyclegan\_nist\_minmax\_lpips\_wavelet & GAN & NIST & 64mT$\to$3T DWI b0 & 1.01 & 25.58 & 0.918 & 0.052 & NC-EPI \\
srcyclegan\_nist\_whitestripe\_balanced\_image & GAN & NIST & 64mT$\to$3T DWI b0 & 1.94 & 27.59 & 0.880 & 0.077 & NC-EPI \\
vitfuser\_nist\_direct & ViT & NIST & 64mT$\to$3T DWI b0 & 0.97 & 27.68 & 0.768 &  & NC-EPI \\
multicontrast\_zenodo\_t2t1flair & diffusion (multicontrast) & Zenodo T2w & T2+T1+FLAIR$\to$3T T2w & 2.40 & 22.40 & 0.716 &  & NC-ANCHOR \\
disgan\_openneuro\_minmax\_lpips\_wavelet & GAN & OpenNeuro T2w & 64mT$\to$3T T2w & 2.53 & 16.79 & 0.297 &  & NC-XDEV \\
disgan\_openneuro\_minmax\_balanced\_image & GAN & OpenNeuro T2w & 64mT$\to$3T T2w & 2.53 & 17.06 & 0.346 &  & NC-XDEV \\
medsynthesis\_openneuro\_minmax\_lpips\_wavelet & GAN & OpenNeuro T2w & 64mT$\to$3T T2w & 1.78 & 16.98 & 0.336 &  & NC-XDEV \\
medsynthesis\_openneuro\_minmax\_balanced\_image & GAN & OpenNeuro T2w & 64mT$\to$3T T2w & 2.00 & 17.14 & 0.373 &  & NC-XDEV \\
srcyclegan\_openneuro\_minmax\_lpips\_wavelet & GAN & OpenNeuro T2w & 64mT$\to$3T T2w & 2.53 & 15.14 & 0.272 &  & NC-XDEV \\
srcyclegan\_openneuro\_minmax\_balanced\_image & GAN & OpenNeuro T2w & 64mT$\to$3T T2w & 2.40 & 15.31 & 0.282 &  & NC-XDEV \\
ressrdiff\_zenodo\_zscore\_fg\_t1w & diffusion & Zenodo T1w & 64mT$\to$3T T1w & 2.53 & 11.94 & 0.339 &  & 16 / 24 (sub-0048 / 0064) \\
ressrdiff\_zenodo\_whitestripe\_t1w & diffusion & Zenodo T1w & 64mT$\to$3T T1w & 2.41 & 10.48 & 0.334 &  & 10 / 24 (5 samples) \\
ressrdiff\_openneuro\_zscore\_fg\_t1w & diffusion & OpenNeuro T1w & 64mT$\to$3T T1w & 1.39 & 12.17 & 0.746 &  & 6 / 12 (HYPE21 / HYPE22) \\
ressrdiff\_openneuro\_whitestripe\_t1w & diffusion & OpenNeuro T1w & 64mT$\to$3T T1w & 1.36 & 11.44 & 0.719 &  & 6 / 16 (5 samples) \\
\midrule
synthsr\_v10\_standard & external (public) & Zenodo + OpenNeuro T1w & 64mT$\to$1 mm T1w & 5.2--6.1 & 7.1--11.0 & 0.07--0.17 &  & 12 / 24 / 6 / 10 \\
\midrule
MEDIAN GAN ($n=18$) & GAN & - & - & 2.06 & - & - & - & - \\
MEDIAN ViT ($n=1$) & ViT & - & - & 0.97 & - & - & - & - \\
MEDIAN diffusion multicontrast ($n=1$) & diffusion & - & - & 2.40 & - & - & - & - \\
MEDIAN diffusion single-contrast ($n=4$) & diffusion & - & - & 1.90 & - & - & - & - \\
\bottomrule
\end{tabular}
}
\end{sidewaystable}

\FloatBarrier
\section{Additional Figures}
\label{app:figures}
\begin{figure}[H]
\centering
\figslot[0.58\linewidth]{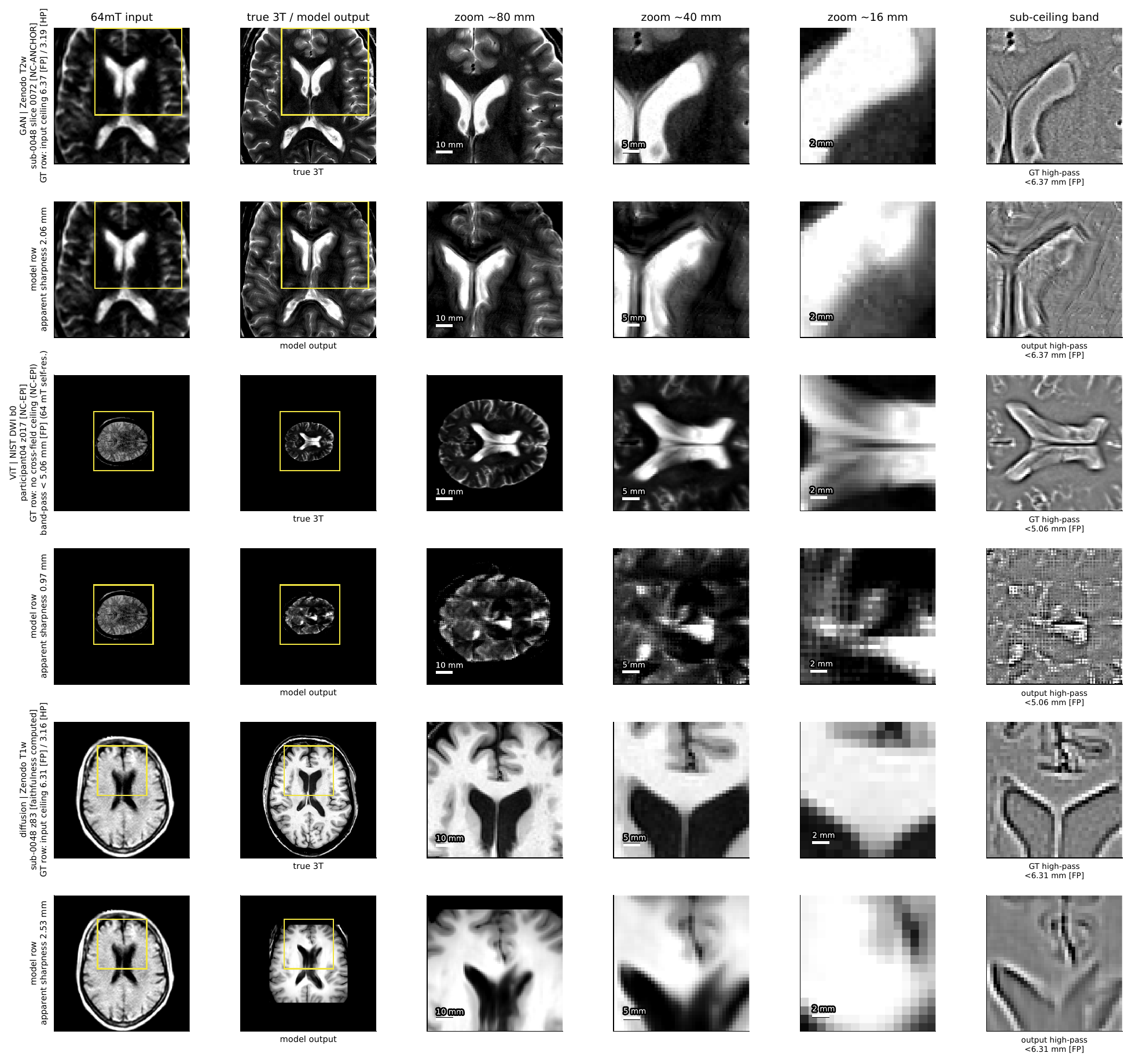}
\caption{\capFINFO}
\label{fig:finfo}
\end{figure}

\begin{figure}[H]
\centering
\figslot[0.72\linewidth]{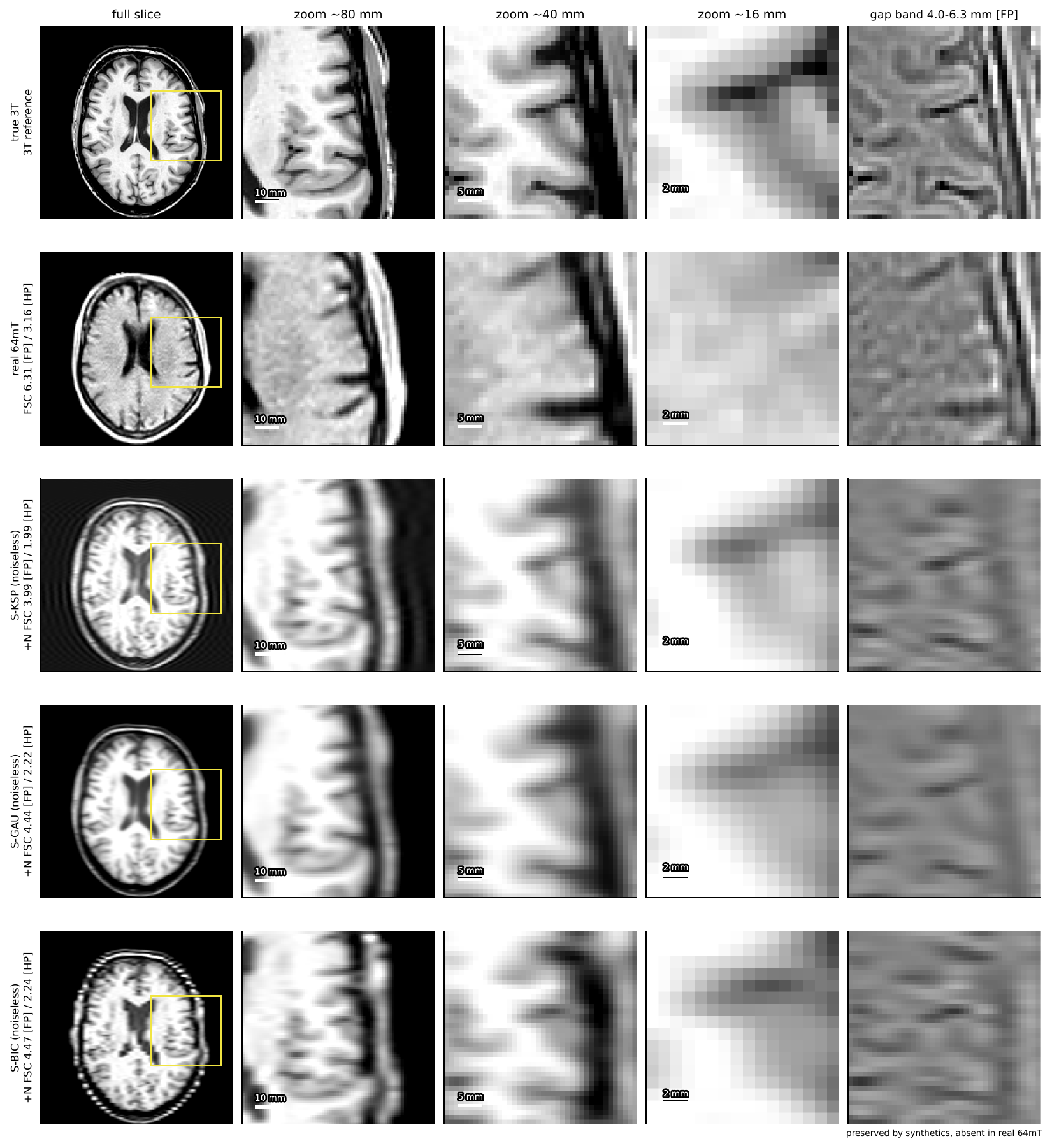}
\caption{\capFSYNTH}
\label{fig:fsynth}
\end{figure}

\begin{figure}[H]
\centering
\figslot[0.58\linewidth]{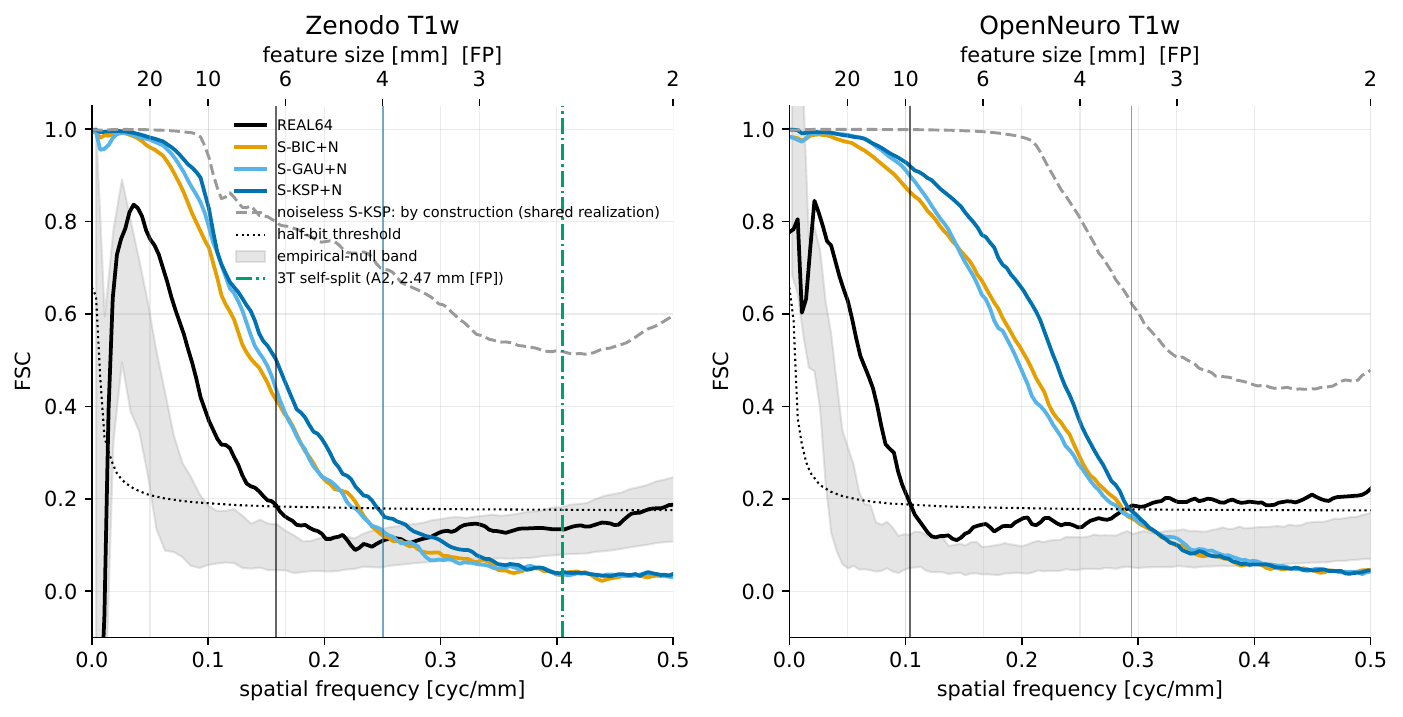}
\caption{\capFEone}
\label{fig:protocol}
\end{figure}

\begin{figure}[H]
\centering
\begin{minipage}[t]{0.40\linewidth}\centering\figslot{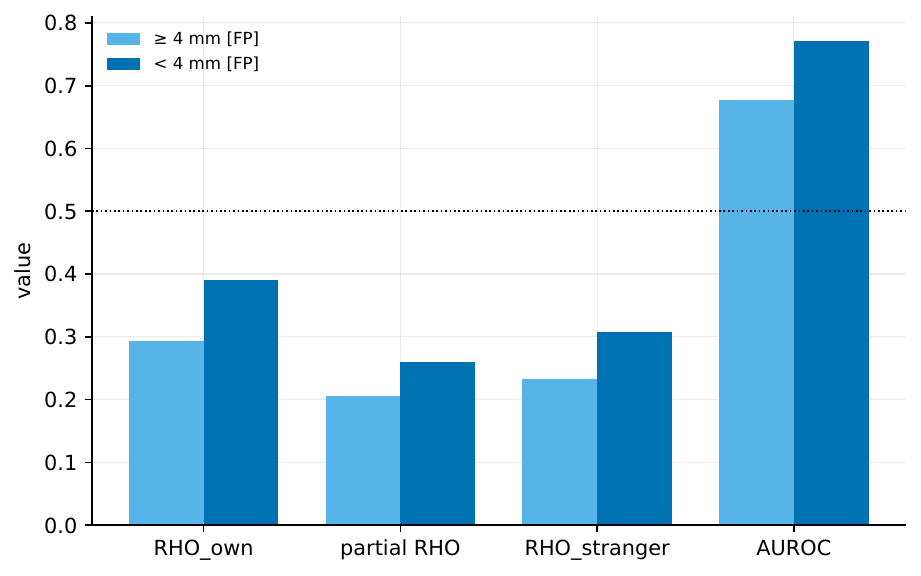}\end{minipage}\hfill
\begin{minipage}[t]{0.58\linewidth}\centering\figslot{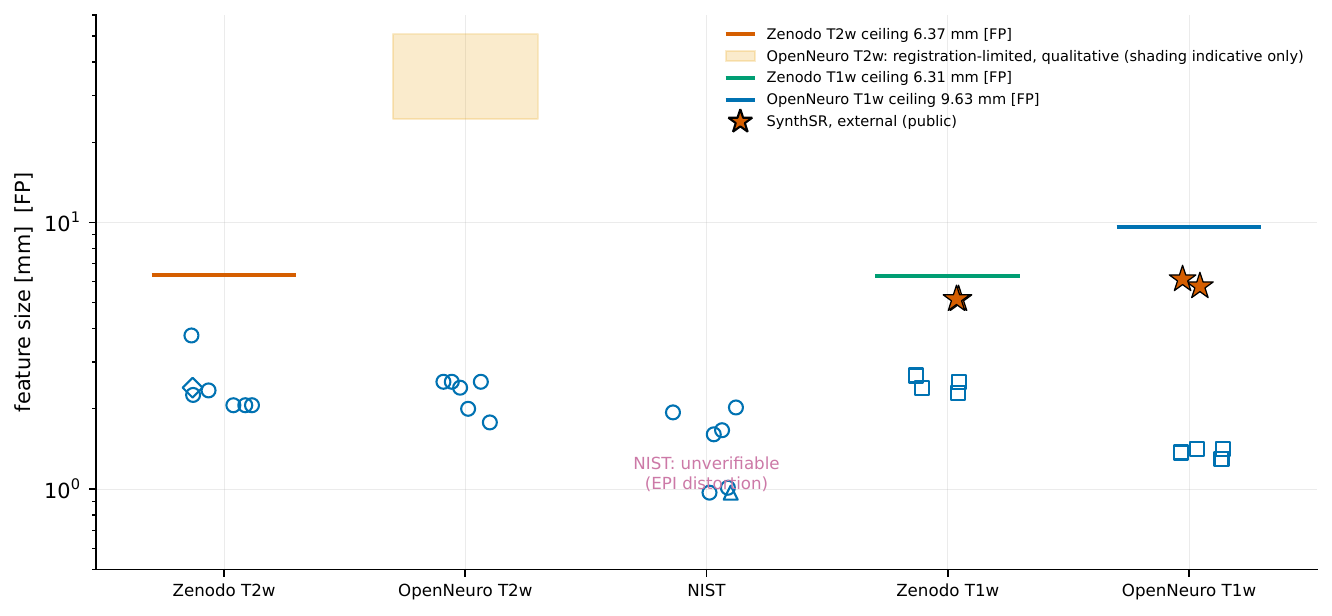}\end{minipage}
\caption{Left: \capFUQ{} Right: \capFSHARP}
\label{fig:fsharp}
\end{figure}

\begin{figure}[H]
\centering
\figslot[0.7\linewidth]{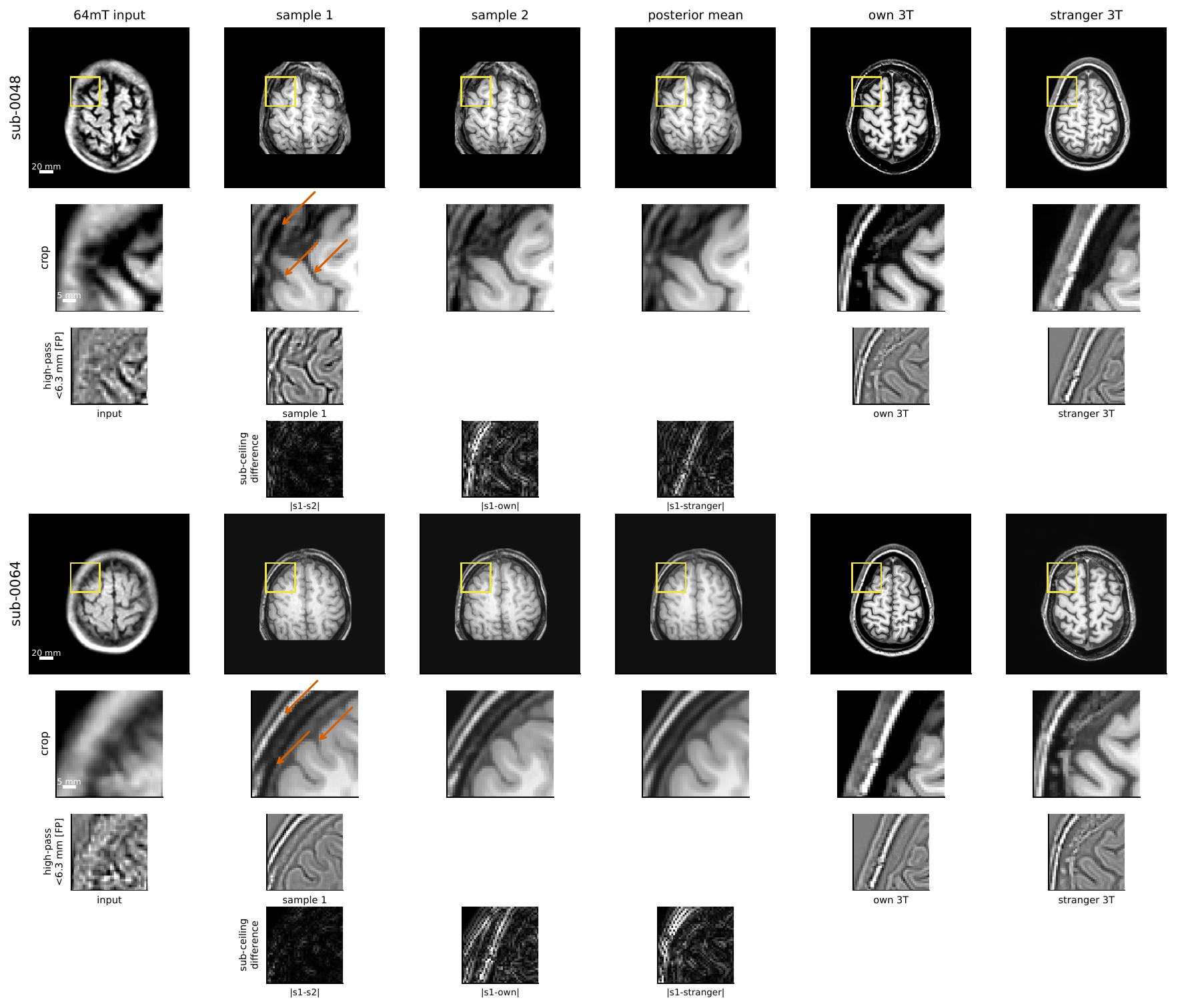}
\caption{\textbf{Fabrication in image space (display only).} Zenodo T1w held-out subjects sub-0048 and
sub-0064: 64 mT input, two posterior samples, posterior mean, own 3T and stranger 3T, with zooms and a high-pass
strip finer than the ceiling. Quantities for these subjects are in Table~\ref{tab:ladder} and
Figure~\ref{fig:fp4}.}
\label{fig:fquala}
\end{figure}

\begin{figure}[H]
\centering
\figslot[0.62\linewidth]{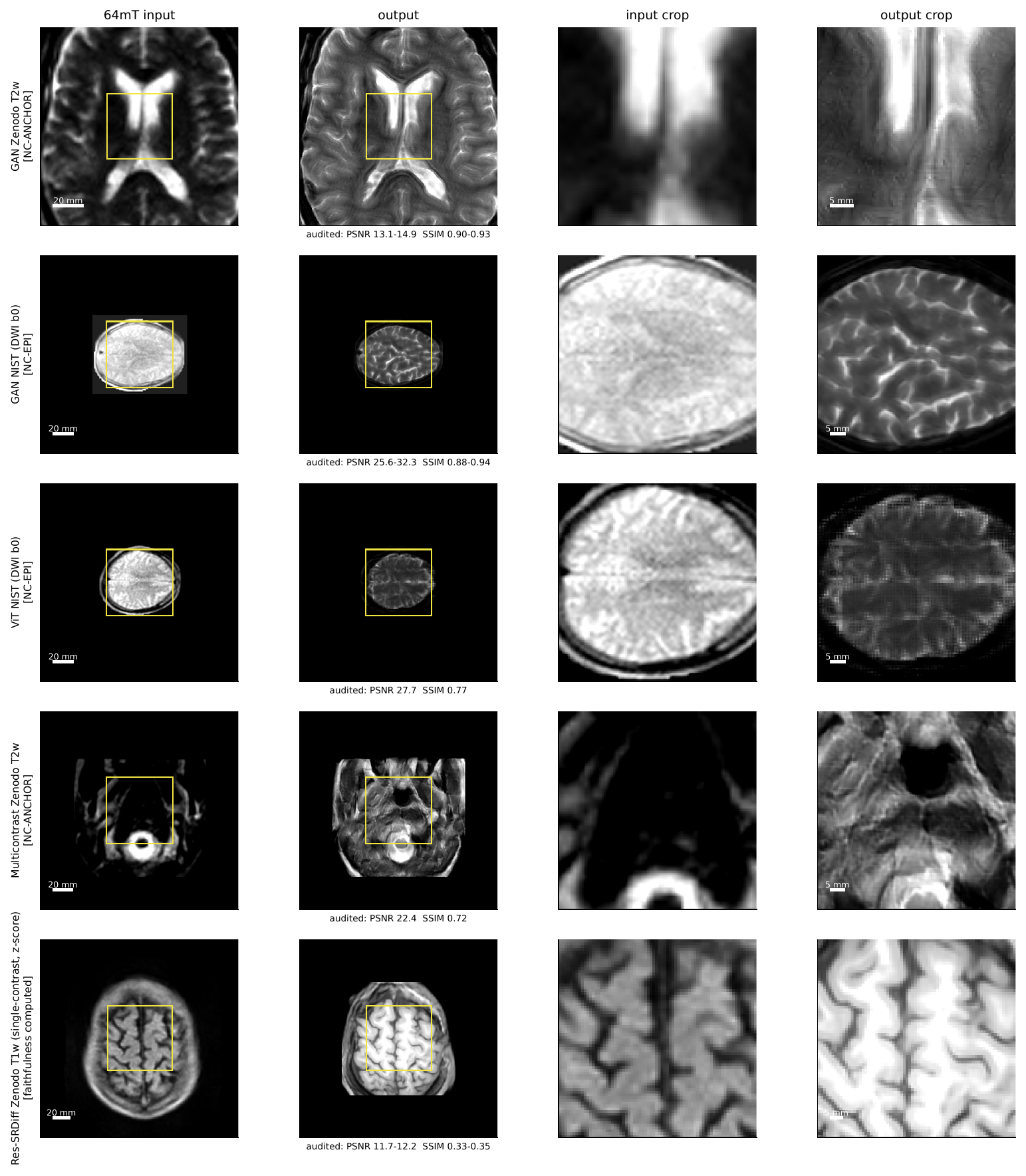}
\caption{\capFQUALB}
\label{fig:fqualb}
\end{figure}

Figure~\ref{fig:finfo} shows the ceiling in image space, Figure~\ref{fig:fsynth} the synthetic-gap band,
Figure~\ref{fig:protocol} the cross-FSC curves, Figure~\ref{fig:fsharp} the uncertainty controls and apparent
sharpness, and Figures~\ref{fig:fquala} and~\ref{fig:fqualb} qualitative fabrication and breadth.

\section{OpenNeuro T2w Ceiling}
\label{app:e2a}
The OpenNeuro T2w recoverability ceiling is registration-limited, qualitative: the exact millimetre value is
uncertain (weak cross-acquisition registration, MI median around $-0.45$, and the reading is beyond
the FSC-validated range), while the qualitative conclusion is robust (the low-field T2w input carries essentially
no faithful fine information about the high-field T2w).

\section{Lessons for Practice: What Failed and Why}
\label{app:practice}

\textbf{Resolution estimators that read the grid instead of the image.} Radial power-spectrum resolution and
slice-accumulated 2D FSC both return the grid limit rather than an image property: accumulating in-plane over
$\sim$149 slices inflates the per-shell voxel count, the half-bit threshold floors, and the
crossing lands at the grid limit (0.69~mm). All audit FSC therefore uses the gated 3D
Algorithm~\ref{alg:fsc}; the FSC null-floor crossing ($\sim$36~mm~\FP{} median, highly variable) is
excluded as a ceiling estimate.

\textbf{Cross-field correlation needs geometric correspondence.} NIST cross-FSC is degenerate
($\sim$65~mm) because the low-field diffusion b0 is EPI-distorted and rigid registration cannot align
it; deformable registration is not allowed, so NIST is reported through self-1FRC and its runs carry NC-EPI.

\textbf{A fixed test split can hide a data pathology from PSNR and SSIM.} The fixed two-subject held-out T2w split
of the Zenodo reconstruction study coincides with the two extreme low-field-resolution outliers of the
distribution, one registration-degenerate and one anomalously fine. PSNR and SSIM on that split do not detect
this; the input anchor (G-M2) does, and it is why those runs carry NC-ANCHOR.

\textbf{A dataset-level gate needs a shared frame.} The OpenNeuro T1w ceiling and its CI are measured on the E1
frame, on which the held-out inputs do not register (anchors 51.4 and
55.8~mm~\FP{}, no crossing under the gate recipe), while on the held-out subjects'
validated v3 frame the dataset ceiling itself is degenerate (median 158~mm~\FP{}, CI
22--512, 12 subjects). Evaluated across frames the gate passes
sub-HYPE21 and fails sub-HYPE22; like against like on E1 both fail and on v3 the outcome reverses
(Table~\ref{tab:frames}). No pairing passes both subjects, so the per-subject anchor on the validated frame is the
operative check on OpenNeuro, and a dataset-level gate should be reported as undefined when the dataset ceiling
and the held-out anchors need different frames.

\textbf{Registration caches must be validated per subject.} The first fabrication analysis ran on a registration
cache whose held-out alignment was worse than the validated one; the raw-input control floored everywhere and
exposed it. Re-running on per-subject validated frames moved the merges to the values reported here. On
OpenNeuro the ordering of the two caches is reversed, so each dataset is analysed on its better-registered frame.

\section{Reconstruction Experiment Details}
\label{app:reconstruction-details}

\textbf{T1w single-contrast diffusion (direct measurement).} Res-SRDiff \citep{safari2025ressrdiff}
(residual-shifting diffusion, exponential schedule with power 0.3, $\kappa=2.0$,
$x_0$ prediction) is trained on 2D axial slices ($256\times256$ at
0.49~mm, padded to 384) of paired scans, rigidly registered on Zenodo and
affinely on OpenNeuro. Zenodo uses 8 training subjects (1220
slices) and OpenNeuro 21; the held-out subjects are sub-0048/sub-0064 and
sub-HYPE21/sub-HYPE22. Intensities are foreground z-scores clipped to $[-3,3]$ or WhiteStripe
white-matter-peak units clipped to $[-10,8]$, both mapped to $[-1,1]$. The loss is MSE and LPIPS
weighted 4:2 plus a gradient term of weight 3; learning rate
$5\times10^{-5}$ with cosine decay, EMA 0.999, batch 1, mixed
precision, no augmentation, validation every 5 epochs with early-stopping patience
5. The diffusion chain has 15 steps in Zenodo training and
4 in OpenNeuro training; all posterior sampling uses 4 steps (seeds
70001--70050 for the z-score arm, five fixed seeds for WhiteStripe). There is no separate
validation split at this sample size: the checkpoint with the best validation PSNR on the two held-out subjects
of each dataset was kept, namely epoch 90 of 100 (Zenodo z-score),
100 (Zenodo WhiteStripe), 40 of 45 (OpenNeuro
z-score), and 5 of a short 6-epoch run (OpenNeuro WhiteStripe, reported
as a robustness check only). PSNR for these runs is computed on foreground voxels after per-volume
0.5/99.5-percentile scaling of prediction and truth to $[0,1]$. On OpenNeuro only a ventricle band of slices is
sampled (92 and 98 slices for sub-HYPE21 and sub-HYPE22) and the remaining slices are filled
from the 3T reference; these fill slices are excluded from every readout (a slice counts as sampled when its
across-sample standard deviation exceeds 15\% of the volume maximum on at least
50 voxels). Algorithm~\ref{alg:rec} and the
uncertainty controls use only the sampled slices, and the apparent sharpness, PSNR, and SSIM of the OpenNeuro
rows in Table~\ref{tab:t2} are computed on the sampled slices only. On Zenodo the model predictions cover a central
125~mm of the field of view, so in the validated frame they cover
75 and 80\% of the brain mask of sub-0048 and sub-0064. Every Algorithm~\ref{alg:rec} readout for
these runs, and for the raw-input benchmark, oracle, WhiteStripe runs, and SynthSR compared against them, is
restricted to the 3T foreground mask (intensity above 5\% of the maximum) intersected with this
coverage and with the sampled slices; only the whole-brain FSC anchor uses the SynthSeg brain mask. After a least-squares linear intensity fit
of the z-score posterior mean to the truth in the same scaled space, PSNR is 13.8--14.9~dB, so
the low values reflect disagreement with the subject rather than intensity scaling.

\textbf{T2w and NIST reference runs.} All reconstruction experiments use acquired paired low-field inputs; no
low-field training input is synthesized. NIST uses paired nominal $b=0$ diffusion trace images (recorded b-values
of approximately $0.05~\mathrm{s/mm^2}$ at 64 mT and $0~\mathrm{s/mm^2}$
at 3T) from 14 training and 3 test participants; OpenNeuro uses paired
T2w from 16 and 4; Zenodo uses paired T2w from 8 and
2. Participants are partitioned before slice extraction. All models are 2D. NIST low-field
images are interpolated to a 1.2~mm in-plane grid and center-padded to
$384\times384$; OpenNeuro and Zenodo low-field volumes are affine-registered to their 3T
references and prepared as $256\times256$ axial slices. All networks are trained from fresh
initialization for 100 epochs. Two training configurations are used. \emph{Balanced} weights the
pixel, LPIPS, and gradient terms $4:2:3$; \emph{perceptual-dominant} weights them
$1:10:1$ and adds a Haar-style consistency term (weight 1: an average-pooled
low-pass L1, plus a high-pass residual L1 for DISGAN). The pixel term is L1 for all GANs and MSE for Res-SRDiff.
All GAN generators predict in the image domain and keep their native adversarial objectives (DISGAN: a
relativistic loss with a Haar wavelet discriminator in both configurations; medSynthesis: WGAN-GP with its
gradient-difference loss; SR-CycleGAN: cycle and identity terms, trained on pairs); only the perceptual-dominant
Res-SRDiff predicts Haar wavelet coefficients. On Zenodo and OpenNeuro both configurations use per-volume
0.5/99.5-percentile min--max scaling (run identifiers ending in \texttt{minmax\_balanced\_image} were originally
named \texttt{whitestripe\_balanced\_image}); on NIST the balanced runs use a WhiteStripe-inspired scaling
\citep{shinohara2014,reinhold2019} and the perceptual-dominant runs slice min--max.

\textbf{Runs outside the audit.} Table~\ref{tab:reconstruction-experiments} marks which trained runs are
audited. The perceptual-dominant Res-SRDiff T2w runs (Zenodo, OpenNeuro), the Res-SRDiff NIST runs, and the
two-input multicontrast runs have held-out slice predictions but were not included in the audit; the balanced
Res-SRDiff T2w runs did not complete training. We list them so that the coverage of the audit is explicit; their
held-out subjects are those of the corresponding audited GAN runs, so faithfulness falls under the same reason
codes, and their sharpness and metric rows are not reported here.

\begin{sidewaystable}
\centering
\small
\caption{Reconstruction experiments of the reference arm. All use acquired paired low-field inputs and
high-field targets. Balanced: pixel/LPIPS/gradient $4:2:3$; perceptual-dominant:
$1:10:1$ plus consistency. Pixel term L1 (GANs) or MSE (Res-SRDiff). Train/Test gives the
participant split; the last column marks runs in the audit (Table~\ref{tab:t2}). The four single-contrast T1w
diffusion runs of the direct measurement are listed first in their dataset blocks.}
\label{tab:reconstruction-experiments}
\resizebox{\linewidth}{!}{%
\begin{tabular}{llllllll}
\toprule
Dataset & Low-field input & High-field target & Model & Normalization & Configuration & Train/Test & Audited \\
\midrule
NIST & Lowest-$b$ trace & Lowest-$b$ trace & Res-SRDiff & Slice min--max & Perceptual-dominant, wavelet prediction & 14/3 & no \\
NIST & Lowest-$b$ trace & Lowest-$b$ trace & Res-SRDiff & WhiteStripe-inspired & Balanced, image prediction & 14/3 & no \\
NIST & Lowest-$b$ + higher-$b$ trace & Lowest-$b$ trace & Res-SRDiff & Training-derived percentile scaling & Balanced, multi-$b$ conditioning & 14/3 & no \\
NIST & Lowest-$b$ trace & Lowest-$b$ trace & DISGAN & Slice min--max & Perceptual-dominant + consistency & 14/3 & yes \\
NIST & Lowest-$b$ trace & Lowest-$b$ trace & DISGAN & WhiteStripe-inspired & Balanced & 14/3 & yes \\
NIST & Lowest-$b$ trace & Lowest-$b$ trace & medSynthesis & Slice min--max & Perceptual-dominant + consistency & 14/3 & yes \\
NIST & Lowest-$b$ trace & Lowest-$b$ trace & medSynthesis & WhiteStripe-inspired & Balanced & 14/3 & yes \\
NIST & Lowest-$b$ trace & Lowest-$b$ trace & SR-CycleGAN & Slice min--max & Perceptual-dominant + consistency & 14/3 & yes \\
NIST & Lowest-$b$ trace & Lowest-$b$ trace & SR-CycleGAN & WhiteStripe-inspired & Balanced & 14/3 & yes \\
NIST & Lowest-$b$ trace & Lowest-$b$ trace & ViT-Fuser (direct) & Training-derived percentile scaling & SSIM + feature-style loss & 14/3 & yes \\
\midrule
OpenNeuro & T1w & T1w & Res-SRDiff & Foreground z-score & Balanced, image prediction & 21/2 & yes \\
OpenNeuro & T1w & T1w & Res-SRDiff & WhiteStripe & Balanced, image prediction & 21/2 & yes \\
OpenNeuro & T2w & T2w & Res-SRDiff & Volume percentile min--max & Perceptual-dominant, wavelet prediction & 16/4 & no \\
OpenNeuro & T2w & T2w & DISGAN & Volume percentile min--max & Perceptual-dominant + consistency & 16/4 & yes \\
OpenNeuro & T2w & T2w & DISGAN & Volume percentile min--max & Balanced & 16/4 & yes \\
OpenNeuro & T2w & T2w & medSynthesis & Volume percentile min--max & Perceptual-dominant + consistency & 16/4 & yes \\
OpenNeuro & T2w & T2w & medSynthesis & Volume percentile min--max & Balanced & 16/4 & yes \\
OpenNeuro & T2w & T2w & SR-CycleGAN & Volume percentile min--max & Perceptual-dominant + consistency & 16/4 & yes \\
OpenNeuro & T2w & T2w & SR-CycleGAN & Volume percentile min--max & Balanced & 16/4 & yes \\
\midrule
Zenodo & T1w & T1w & Res-SRDiff & Foreground z-score & Balanced, image prediction & 8/2 & yes \\
Zenodo & T1w & T1w & Res-SRDiff & WhiteStripe & Balanced, image prediction & 8/2 & yes \\
Zenodo & T2w & T2w & Res-SRDiff & Volume percentile min--max & Perceptual-dominant, wavelet prediction & 8/2 & no \\
Zenodo & T2w & T2w & DISGAN & Volume percentile min--max & Perceptual-dominant + consistency & 8/2 & yes \\
Zenodo & T2w & T2w & DISGAN & Volume percentile min--max & Balanced & 8/2 & yes \\
Zenodo & T2w & T2w & medSynthesis & Volume percentile min--max & Perceptual-dominant + consistency & 8/2 & yes \\
Zenodo & T2w & T2w & medSynthesis & Volume percentile min--max & Balanced & 8/2 & yes \\
Zenodo & T2w & T2w & SR-CycleGAN & Volume percentile min--max & Perceptual-dominant + consistency & 8/2 & yes \\
Zenodo & T2w & T2w & SR-CycleGAN & Volume percentile min--max & Balanced & 8/2 & yes \\
Zenodo & T2w + T1w & T2w & Res-SRDiff & Training-derived percentile scaling & Balanced, multicontrast conditioning & 8/2 & no \\
Zenodo & T2w + FLAIR & T2w & Res-SRDiff & Training-derived percentile scaling & Balanced, multicontrast conditioning & 8/2 & no \\
Zenodo & T2w + T1w + FLAIR & T2w & Res-SRDiff & Training-derived percentile scaling & Balanced, multicontrast conditioning & 8/2 & yes \\
\bottomrule
\end{tabular}}
\end{sidewaystable}

\section{Reproducibility of the External SynthSR Arm}
\label{app:synthsr}
The external arm uses the publicly released SynthSR \citep{iglesias2021synthsr}, the standard single-input model
\texttt{SynthSR\_v10\_210712.h5} at repository commit \texttt{9e63935}, run with
\texttt{scripts/predict\_command\_line.py -{}-cpu -{}-threads 4} on each held-out subject's registered 64 mT
scan; each output is mapped onto that subject's analysis frame by a single interpolation before the identical
Algorithm~\ref{alg:rec}, apparent-sharpness, and foreground PSNR/SSIM readouts. The environment is Python 3.8.10
with TensorFlow 2.2.0, Keras 2.3.1, and protobuf pinned to 3.19.6 (TensorFlow 2.2 is incompatible with protobuf
4.x). The released commit carries two dictionary-access defects in the inference script (\texttt{args.model} and
\texttt{args.disable\_flipping}, where \texttt{args} is a dictionary built by \texttt{vars(parse\_args())}); we
correct both to item access, changing no weights and no inference behaviour. On sub-0064 the SynthSR output
correlates only 0.10 with the 3T at the coarsest band, below the harness alignment warning
(0.2), so its merge there should be read with care; the same subject is marginal on the G-M3
correspondence check (Appendix~\ref{app:gates}). On sub-HYPE22 its REC is 0.04 and
0.00 against SWAP of $-0.043$ and $-0.058$ at the 16 and
12~mm bands, which is why its unclipped merge (10~mm) is finer than its benchmark. The dual-input low-field variant
\citep{iglesias2023lowfield} is not audited because these held-out subjects lack a paired 64 mT T2w.

\end{document}